\documentclass{new_tlp}
\usepackage{mathptmx}

\newcommand{\svadalog}{\textsc{soft vadalog}\xspace}
\newcommand{\vadalog}{\textsc{vadalog}\xspace}

\usepackage{graphicx}

\usepackage{algorithm}
\usepackage{xspace}
\usepackage{url}

\usepackage[noend]{algpseudocode}
\usepackage{type1cm}        
\usepackage{graphicx}        
\usepackage[bottom]{footmisc}
\usepackage{newtxmath}       

\usepackage{changebar}
\newcommand{\change}[1]{#1}

\newcommand{\changeR}[1]{#1}

\usepackage[dvipsnames]{xcolor}
\definecolor{bittersweet}{rgb}{1.0, 0.44, 0.37}
\definecolor{emerald}{rgb}{0.31, 0.78, 0.47}

\newtheorem{example}{Example}[section]
\newtheorem{proposition}{Proposition}

\usepackage{enumitem}
\usepackage[T1]{fontenc}
\usepackage{xspace}
\usepackage{wrapfig}
\usepackage{comment}

  \title[Theory and Practice of Logic Programming]
        {\change{Swift Markov Logic for\\ Probabilistic Reasoning on Knowledge Graphs}}

  \author[L. Bellomarini, E. Laurenza, E. Sallinger and E. Sherkhonov]
         {Luigi Bellomarini, Eleonora Laurenza\\
         Banca d'Italia
         \and Emanuel Sallinger\\
         TU Wien and University of Oxford
         \and Evgeny Sherkhonov \\
         University of Oxford
         }

\pagerange{\pageref{firstpage}--\pageref{lastpage}}

\begin{document}

\label{firstpage}

\maketitle

\begin{abstract}
We provide a framework for probabilistic reasoning in Vadalog-based Knowledge Graphs (KGs), satisfying the requirements of ontological reasoning: full recursion, powerful existential quantification, expression of inductive definitions.
Vadalog is a \change{Knowledge Representation and Reasoning} (KRR) language based on Warded Datalog+/-, a logical core language of existential rules, with a good balance between computational complexity and expressive power.
Handling uncertainty is essential for reasoning with KGs. Yet Vadalog and Warded Datalog+/- are not covered by the existing probabilistic logic programming and statistical relational learning approaches for several reasons, including insufficient support for recursion with existential quantification, and the impossibility to express inductive definitions.
In this work, we introduce Soft Vadalog, a probabilistic extension to Vadalog, satisfying these desiderata. 
\change{A Soft Vadalog program induces what we call a Probabilistic
Knowledge Graph (PKG), which consists of a probability distribution on
a network of chase instances, structures obtained by grounding the
rules over a database using the chase procedure.}
We exploit PKGs for probabilistic marginal inference. We discuss the theory and present MCMC-chase, a Monte Carlo method to \change{use} Soft Vadalog in practice. We apply our framework to solve data management and industrial problems, and experimentally evaluate it in the Vadalog system.\\ 
\smallskip\noindent
Under consideration in Theory and Practice of Logic Programming (TPLP).
\end{abstract}

  \begin{keywords}
    Knowledge Graphs  \and Reasoning \and Datalog+/- \and Markov Logic Networks.
  \end{keywords}


\section{Introduction}
\label{sec:introduction}
Knowledge Representation and Reasoning (KRR) languages based on logic rules are experiencing a significant resurgence in the context of Knowledge Graph (KG) systems. Alongside the graph-based data model of property graphs~\cite{Angl18}, more and more companies and users yearn for the possibility to harness domain knowledge in the \textit{intensional} components of the graph~\cite{BFGS19} and be able to use it to solve complex reasoning tasks, well beyond the initial applications of ontological reasoning of the early times.

KRR languages play a fundamental role in this, and, in order to cope with complex real-life applications, should support a number of desiderata. First, \change{the knowledge management community has been showing a strong appreciation for rule-based languages, wishing to benefit from} 
all the typical advantages over procedural approaches, such as user-orientation, modularity and, of course, explainability. Then, KRR languages should be syntactically simple and achieve high expressive power in order to be able to deal with complex industrial use cases. At the same time, they should guarantee low data complexity (when the set of rules is fixed and the data vary), so to be executable in practice and scale with large volumes of data~\cite{BGPS17}.

\medskip
\noindent
{\sc vadalog} is a logic KRR language based on Warded Datalog$^\pm$~\cite{BeSaGo18},
a member of the Datalog$^\pm$ family~\cite{CaGP12}.
Datalog$^\pm$ languages generalize Datalog by adopting \textit{existential rules} 
with existential quantification in the rule head. The presence of existential quantification and recursion makes reasoning undecidable in general, as infinitely many symbols may be generated in the logical entailment~\cite{GoPi15}. Warded Datalog$^\pm$ introduces  syntactic restrictions that limit the propagation of nulls while preserving high expressive power.

\medskip\noindent
Let us start with the following example to introduce the reasoning task.

\begin{example}
\label{ex:first}
\emph{Consider a Knowledge Graph $G$. The facts in its extensional component describe domain relationships between constants $a, b, c, l, m, n$ as follows:}
\change{\begin{align*}
\{&{\rm Contract}(a,b,c), {\rm Exposure}(b,l), {\rm RegulatoryRestriction}(m,l), {\rm LenderClass}(m,n)\}
\end{align*}}
\noindent
\emph{Let us extend $G$ with the following Warded Datalog$^\pm$ existential rules (the intensional component of the KG), \change{encoding a portion of the credit domain, from our industrial use cases:}}
\change{
\begin{align*}
{{\rm LenderType}(x,y)}, {\rm RegulatoryRestriction}(y,z) \rightarrow &~\exists v \, {\rm Guarantee}(x,z,v) & (1) \\
{{\rm LenderType}(x,y)}, {\rm LenderClass}(y,z) \rightarrow &~{\rm LenderType}(x,z) & (2) \\
{{\rm Contract}(x,y,z)}, {\rm Exposure}(y,w) \rightarrow &~{\rm Contract}(z,w,x) & (3)\\
{{\rm Contract}(x,y,z)}, {\rm RegulatoryRestriction}(w,y) \rightarrow &~{\rm LenderType}(x,w). & (4) 
\end{align*}}
\noindent\changeR{
\emph{Rule~(1) encodes that if a lender $x$ is of type $y$ (e.g., a bank, a small company, etc.) and \changeR{lenders of type $y$ are subject to \textsf{RegulatoryRestrictions}, enforced by financial supervision authorities, that require loans to be covered by a specific \textsf{Guarantee} (known as ``collateral'') of type $z$ (e.g., securities, real estate properties, cash, etc.), then, such \textsf{Guarantee} for lender $x$ exists and involves a guarantor $v$, the entity issuing the guarantee itself, be it an individual, a bank, or a financial intermediary.} 
\textsf{LenderClass} in Rule~(2) defines a taxonomy of lenders, classifying them into classes (e.g., mortgage lenders, retail lenders, direct lenders, etc.), so that if $y$ is a subclass of $z$ (e.g., credit unions is a subclass of retail lender) and $x$ is of type $y$, it follows that $x$ is of type $z$ as well. In practice, loans are formalized by \textsf{Contracts}: in the body of Rule~(3), $x$ is the lender, $z$ the borrower, and $y$ the type of contract, encoding the type of loan. Different types of loans give rise to different forms $w$ of financial \textsf{Exposure}, i.e., the type of repayment obligation that $z$ has towards $x$. Rule~(3) justifies the existence of another \textsf{Contract}, from $z$ to $x$, witnessing such repayment obligation encoded by the contract type $w$ associated to the exposure. Finally, Rule~(4) complements Rule~(2), so if a \textsf{Contract} from $x$ to $z$ is in place to satisfy a restriction that requires a guarantee encoded by the contract type $y$, based on a lender type $w$, then $x$ is of type $w$.}}

\hfill$\blacksquare$
\end{example}

\noindent An example of (ontological) reasoning task over $G$ corresponds to answering the query: ``What are all the $\textsf{Contracts}$ and $\textsf{Guarantees}$ that are expected to be entailed by the KG?''.
They are $\textsf{Contract}(c,l,a)$ and $\textsf{Guarantee}(c,l,v_0)$,
where $v_0$ is a fresh arbitrary value (a \emph{labeled null}). 

\smallskip
Let us consider a modified setting for Example~\ref{ex:first}, in which the assertions are not definitive, but hold with different strengths, \change{due to heterogeneous implementations of the regulations by the different financial intermediaries.} 
We prefix our rules with a weight proportional to the bias we have for them to hold (indicated by the number before the :: symbol) or, under another perspective, that depends on \change{how often the rules are satisfied in the available example data sets.} 

\begin{example}
\label{ex:second}\textit{We extend $G$ of Example~\ref{ex:first} with the following set of weight-prefixed Warded Datalog$^\pm$ rules.}
\change{
\begin{align*}
0.9:: {{\rm LenderType}(x,y)}, {\rm RegulatoryRestriction}(y,z) \rightarrow &~\exists v \, {\rm Guarantee}(x,z,v) & (5) \\
0.8 :: {{\rm LenderType}(x,y)}, {\rm LenderClass}(y,z) \rightarrow &~{\rm LenderType}(x,z) & (6) \\
0.7 :: {{\rm Contract}(x,y,z)}, {\rm Exposure}(y,w) \rightarrow &~{\rm Contract}(z,w,x) & (7)\\
{{\rm Contract}(x,y,z)}, {\rm RegulatoryRestriction}(w,y) \rightarrow &~{\rm LenderType}(x,w). & (8) 
\end{align*}}
\textit{\change{We represent here some notion of uncertainty related to our domain: depending on how each financial intermediary implements the regulations, some rules may apply or not. In particular, Rules~5-7 tend to be respected more or less regularly as reflected by their weights, whereas Rule~8 is a hard rule}~$\blacksquare$}
\end{example}

\noindent
A probabilistic reasoning task would consist in answering, over logic programs, queries like: \emph{``What is the probability for each $\textsf{Contract}$ and \textsf{Guarantee} to be entailed?'}. We wish
to compute the marginal probability of entailed facts, so, e.g.,  of $\textsf{Contract}(c,l,a)$ and $\textsf{Guarantee}(c,l,v_0)$.

\smallskip
Beyond our running example, many scenarios of practical applications of probabilistic reasoning in Warded Datalog$^\pm$ are common and span many domains we have direct experience on with Datalog$^\pm$, besides the financial domain~\cite{BFGS19}, the automation of data science pipelines~\cite{BFGK18}, data acquisition~\cite{MFLS17}, data extraction~\cite{FSSF18}, and others.

\smallskip
To enable such scenarios, we need KRR languages able to \changeR{perform probabilistic reasoning in the context of the requirements of ontological reasoning}: (i) powerful existential quantification, supporting the quantification of SPARQL and OWL 2 QL, (ii)  \changeR{full recursion, capturing full Datalog,
to express general non-ground inductive definitions (e.g., transitive closure, like in the standard path definition $\textit{edge}(X,Y)\to \textit{path}(X,Y)$;
$\textit{edge}(X,Z), \textit{path}(Z,Y)\to \textit{path}(X,Y)$)~\cite{FiBrRe15,LeWa16}.} 
Although probabilistic reasoning is of interest in four broad research areas, namely \emph{probabilistic logic programming} (e.g., ProbLog)~\cite{ABCR17,RaedtK15,Pool08,Rigu08,Sato95,SatoK97,VeVB04}, \emph{probabilistic programming languages} (e.g., BLOG)~\cite{GMRB08,KeRa08,MilchMRSOK05,Pfef09}, \emph{statistical relational learning} (e.g., Markov Logic Networks)~\cite{Jaeg18,LeWa16,RiDo06} \change{and \textit{probabilistic knowledge bases}~\changeR{\cite{BoCL18,JuLu12}}} as we shall see, none of the existing approaches fit our requirements. In particular they either fail in providing simultaneous support for recursion and existential quantification or do not allow for inductive definitions \changeR{in uncertain ontological reasoning settings.} 

 \smallskip\noindent \textbf{Contribution.} 
In this work we introduce a theoretical framework and workable system that implements probabilistic reasoning on \vadalog KGs. In particular, we contribute the following:

\begin{itemize}[leftmargin=2mm]\itemsep-\parsep
\item{We present $\svadalog$, a probabilistic extension to {\sc vadalog}. This KRR language allows to define \textit{Probabilistic Knowledge Graphs}, that is, a probabilistic version of the notion of \change{a} knowledge base or KG commonly adopted in automated reasoning contexts. In particular, a \svadalog program and a database define a probability distribution over the nodes of a \textit{chase network}, a data structure (technically, a \textit{probabilistic graphical model}) built by the application of the {\sc chase} procedure. Here we combine both the experience from the database community, where chase procedures \change{generate the entailed facts,} 
and \change{the experience} from statistical relational learning, as \svadalog programs are templates for chase networks, along the lines of Markov Logic Networks, where weighted FO formulas generate Markov networks.}
\item{We characterize the problem of reasoning on PKGs and conclude it is \#\texttt{P}-hard.}
\item{Based on the mentioned intractability result, we propose the \emph{MCMC-chase}, an approximate technique for marginal inference  combining a Markov Chain Monte Carlo method (specifically the Metropolis-Hastings algorithm) with a \emph{chase-based} procedure. Chase procedures are used in databases to enforce logic rules by generating entailed facts. Here, the chase procedure is guided by MCMC and marginal inference is performed in the process. We present the details of the algorithm and define and prove its theoretical underpinnings.}
\item{We illustrate applications of probabilistic reasoning to \textit{record linkage}~\cite{Chri12} and \textit{data fusion}~\cite{BlNa08}, two core data management problems.}
\item{Finally, an extension of the {\sc Vadalog} system implementing the algorithm is experimentally evaluated on real-world and synthetic cases in a corporate economics setting.}
\end{itemize}

\noindent
This invited paper is a substantially extended version of a recent short work~\cite{BLSS20}, where we initially introduced \svadalog. In particular, all the sections contain relevant new material and more detailed explanations. The background is broadly developed; theoretical results are fully reported, proven and discussed; 
applications to data management problems and experimental settings are completely new.
 
\medskip\noindent\textbf{Overview}. In Section~\ref{sec:relwork} we provide motivation for our approach and analyze the related work. In Section~\ref{sec:kgs} we introduce Warded Datalog$^\pm$, \vadalog and the needed background. 
In Section~\ref{sec:pkgs}, we introduce PKGs, \svadalog and probabilistic reasoning with them. In Section~\ref{sec:mcmc-chase}, we present the MCMC-chase algorithm and its theoretical underpinnings. 
In Section~\ref{sec:use-cases}, we discuss the application use cases. Experimental settings are in Section~\ref{sec:experiments}. Section~\ref{sec:conclusion} concludes the paper.

\section{Motivation and Related Work}
\label{sec:relwork}
\change{The combination of logic-based approaches and uncertainty in the form of probability has been a long-lasting goal of artificial intelligence. Seminal theoretical works~\cite{Halp89,Bacc90} make a preliminary distinction between \textit{statistical inquiries} on general characteristics of the domain (e.g., in our example, ``what is the probability for any random $\textsf{Contract}$ to be entailed by a matching $\textsf{LenderType}$ and $\textsf{RegulatoryRestriction}$?'') and \textit{degree of belief} (e.g., ``what is the probability that the $\textsf{Contract}(c,l,\nu_0)$ is entailed by $\textsf{LenderType}(c,m)$ and $\textsf{RegulatoryRestriction}(m,l)$ ?'') and provide a framework to combine the two approaches. The former requires accounting for a probability distribution over the domain, the latter over the possible worlds. Along the lines of successful \textit{Statistical Relational Learning} approaches~\cite{SiDo05}, for ontological reasoning on KGs, we are interested in both the perspectives, and need to handle both uncertainty on the rules (and so the degree of belief) and---as a special case of it---on the data (and so the statistical characteristics of the domain). Beyond the early frameworks, which assume a fixed probabilistic distribution over the possible worlds, the complex domains in which KGs are used require handling a broader range of probabilistic relations between the domain objects.} 

\smallskip
In the areas of \emph{Probabilistic Logic Programming} (PLP), \emph{Probabilistic Programming Languages} (PPL), \emph{Statistical Relational Learning} (SRL), \change{and \emph{Probabilistic Knowledge Bases (PKB),}} we find different angles to approaching uncertainty in reasoning \change{to such a broader extent.} All these areas can be considered related to our work, yet none of them singularly satisfies the desiderata we have laid out in the introduction and can be directly used for our purpose of providing a probabilistic extension to \vadalog and Warded Datalog$^\pm$ in particular.

\smallskip
\noindent
\emph{Probabilistic Logic Programming} (PLP) approaches~\cite{RaedtK15} mostly adopt the well-known distribution semantics~\cite{Sato95}. According to these semantics, a program induces a probability distribution over a set of different programs (worlds). The marginal probability of a fact is then obtained as proportional to the number of models in which that fact is true, when finite. On this basis, different variants of the distribution semantics can be characterized and exist, depending on the adopted notion of model for a world, such as the standard ones: minimal stratified model, stable model, well-founded model.


PLP offers insufficient support in handling recursion and existentials together in a single decidable fragment. Apart from PLP languages that do not support recursion, most of the other frameworks and systems do not allow for non-ground recursive probabilistic rules including the creation of new values via existential quantification, or do not disclose details about how it would be handled. This is the case of Probabilistic Logic Programs~\cite{Dant91}, Probabilistic Horn Abduction~\cite{Pool93}, CP-Logic~\cite{Venn09}, ICL~\cite{Pool08}, PRISM~\cite{SatoK97}, LPAD~\cite{Rigu08,VeVB04}, ProbLog~\cite{RaedtK15}, cPlint~\cite{ABCR17}, cProbLog~\cite{LBDK17}.

These limitations are self-evident going back to Example~\ref{ex:second} and trying to answer the query  \emph{``What are all the entailed ${\rm Contracts}$?''} over $G$.  The existing PLP techniques fail to conclude ${\rm Contract}(c,l,v_0)$, because they abort when running into probabilistic recursive rules that involve the creation of new values, like for instance Rule~5 and Rule~8, or Rule~6.


\smallskip
\noindent
\emph{Probabilistic Programming Languages} (PPL) systems and frameworks such as BLOG~\cite{MilchMRSOK05}, BLP~\cite{KeRa08}, Church~\cite{GMRB08}, Figaro~\cite{Pfef09}, are outside our scope of interest. They are in fact typically based on an underlying Bayesian network model~\cite{KoFr09} and forbid recursion or existential quantification by construction.

\smallskip
\noindent
\emph{Statistical Relational Learning} (SRL) approaches have the great merit of pursuing a conciliation of logic-based reasoning and probability. Our yardstick here is \textit{Markov Logic}~\cite{DoLo19} and \emph{Markov Logic Networks (MLN)}~\cite{RiDo06} in particular, which are a probabilistic generalization of both FOL and probabilistic graphical models, such as \emph{Relational Bayesian Networks}~\cite{Jaeg18}. Markov logic relaxes the notion of hard constraints of ordinary logical knowledge bases, and introduces soft constraints. A MLN is composed of a set of rules adorned with weights, expressing the rule \change{relative importance in the domain description.} In other words, FO rules are a template for the definition of Markov networks, where the actual probabilistic inference takes place after the construction of the network (grounding). A model for a MLN need not satisfy all the MLN rules and the likelihood of a model depends on the weight of the rules it satisfies. \changeR{Then, the marginal probability of a fact in turn aggregates the marginal probability of the models in which it holds. 
This is also the case for facts not entailed by the FOL theory. That is, a fact holding only in some models (but not all models, i.e., the fact is not logically entailed) can have non-zero probability.}
Unfortunately, \changeR{in the ontological reasoning context, this behaviour results in the major limitation that 
 MLNs cannot enforce marginal probability zero for models having facts not in the range of the transitive closure.} This is just one example of a broader area that refers to the ``ability to express (non-ground) inductive definitions''~\cite{FiBrRe15}, such as a graph path in terms of its edges. Example~\ref{ex:second} immediately shows \changeR{how this can lead to potentially incorrect results.} For example, \change{under the closed-world assumption (CWA) used in MLNs,} as $\{\textsf{Contract}(a,b,c),\textsf{Exposure}(b,n)\}$ falsifies the premise of Rule~7 \change{($\textsf{Exposure}(b,n) \notin D$ and is therefore false by CWA), the rule does not require the presence of the \changeR{fact} $\textsf{Contract}(c,n,a)$ in models. Yet, \changeR{it is impossible to enforce probability zero for} models containing $\textsf{Contract}(c,n,a)$, or $\textsf{Contract}(c,b,a)$, or any other fact consistent with the theory.
}    

MLNs are therefore unsuitable to be directly adopted for ontological reasoning, where constructive inductive definitions are understood as a natural form of representation of human knowledge. 
In the context of KG reasoning, the requirement for general inductive definitions is even stronger, as nontrivial navigation of graph-based structures and individuation of edge patterns are core applications.
\changeR{For example, consider again the standard path definition, where rules hold with different strengths, e.g., for data quality reasons: $0.9 :: \textit{edge}(X,Y)\to \textit{path}(X,Y)$;
0.8 :: $\textit{edge}(X,Z), \textit{path}(Z,Y)\to \textit{path}(X,Y)$. Given the database $D=\{\textit{edge}(a,b),\textit{edge}(c,d)\}$, it would be impossible to assign probability zero to the fact \textit{path}(a,d), expressing a non-existing path.}

To complete the overview of SRL, it is worth mentioning two more proposals for probabilistic logical reasoning that also draw inspiration from Markov logic. One is
$\text{LP}^{\text{MLN}}$, a relevant logic programming approach~\cite{LeWa16} largely based on MLNs. Unfortunately, it is unsuitable for our purposes as it relies on stable model semantics instead of well-founded or simply stratified semantics, which we consider the standard option. \change{The other has appeared in the context of Datalog$^\pm$ ontologies~\cite{GLVS13},} yet has a specific focus on database repair and, more importantly, strict adherence to Markov logic semantics. We also draw inspiration from these approaches, but consider the KG reasoning setting and the consequent desiderata, which make Markov logic not directly applicable for the reasons we have seen in this section.

\smallskip
\noindent
\change{\emph{Probabilistic Knowledge Bases} (PKB). Querying large-scale probabilistic knowledge bases is an interesting perspective of uncertain reasoning, with a broad range of proposals~\cite{BoCL18}. Those based on initial models of \textit{probabilistic databases}~\cite{DaSu07} assume the probabilistic independence of the facts, while still encountering major computational challenges as captured by the notorious \textit{dichotomy result} for the evaluation of conjunctive queries~\cite{DaSu12}.}
\change{Currently, more advanced approaches stemming from probabilistic databases basically allow to model any probability distribution over the set of possible worlds~\cite{GeTa06}. Recognizing that probabilistic query evaluation is closely connected to the problem of weighted first-order model counting~\cite{BBGS14}, \textit{knowledge compilation}, \textit{database factorization}, \textit{tensor factorization} techniques have been proposed~\cite{Olte16,OlSc16,KrNi14} to cope with computational complexity and, in this line, a number of \textit{reasoning systems} based on approximate query answering arose, with remarkable examples such as SlimShot~\cite{GrSu16}, MayBMS~\cite{HAKO09}, Tuffy~\cite{NRDS11}, and others as summarized in a recent survey~\cite{BrSu17}.}
\change{This entire line of research shares with our work the need for representing increasingly complex probabilistic relations between domain entities, yet does not immediately lead to frameworks \changeR{that are directly applicable for ontological reasoning with uncertainty on knowledge graphs, because they adopt the closed-domain assumption, usual in the database context.}}

\change{Removing the closed-domain assumption, uncertain reasoning has been studied in the context of \textit{ontology-based access (OBDA) to probabilistic data}~\cite{PLCD08} with both \textit{lightweight description logic}~\changeR{\cite{JuLu12}} and Datalog$^\pm$ formalisms~\cite{BoCL17}. For the first category, relevant works~\cite{CePe15,DaFL08} combine the $\mathcal{EL}$ and \textit{DL-Lite} families of description logics and Bayesian networks. Yet, while admitting recursion and existential quantification, such ontological languages do not offer support for multi-attributed structures, for example those needed in property graphs even for basic reasoning tasks such as symmetric relations, and are thus unsuitable for KGs~\cite{MaKr17}. The second category connects to the frameworks we have seen in the SRL area, such as those combining Datalog$^\pm$ and Markov logic~\cite{GLVS13}, with the explored limitations for transitive closure.}

\section{Preliminaries}
\label{sec:kgs}\label{sec:kgs_action}

This section recalls the preliminary notions of Warded Datalog$^\pm$. 

\medskip
\noindent
Let $\mathbf{C}$, $\mathbf{N}$, and $\mathbf{V}$ be disjoint countably infinite sets of {\em constants}, {\em (labeled) nulls} and (regular) {\em variables}, respectively. A {\em (relational) schema} $\mathbf{S}$ is a finite set of relation symbols (or predicates) with associated arity. A {\em term} is a either a constant or variable. An {\em atom} over $\mathbf{S}$ is an expression  $R(\bar v)$, where $R \in \mathbf{S}$ is of arity $n > 0$ and $\bar v$ is an $n$-uple of terms. 
A {\em database instance} (or simply \textit{database}) over $\mathbf{S}$ associates to each relation symbol in $\mathbf{S}$ a relation of the respective arity over the domain of constants and nulls. The members of relations are called \textit{tuples}. 

Warded Datalog$^\pm$, a member of the Datalog$^\pm$ family~\cite{CaGP12}, extends Datalog~\cite{ceri1989you} with existential quantification to support ontological reasoning (whence the $+$ symbol in $\pm$) and stratified negation, while at the same time, restricts other aspects of the syntax in order to guarantee \textit{decidability} and \textit{tractability} of the reasoning task (the $-$ symbol in $\pm$). A \emph{rule} is a first-order sentence of the form \(
\forall \bar x \forall \bar y (\varphi(\bar x,\bar y)\ \rightarrow\ \exists \bar z \, \psi(\bar x, \bar z))
\), 
where $\varphi$ (the {\em body}) and $\psi$ (the {\em head}) are conjunctions of atoms. We omit universal quantifiers and use comma to denote conjunctions, as usual in this context. A \textit{program} is a set of rules.  \change{Towards introducing the semantics of Datalog$^\pm$, let us ignore existential quantification for the moment, and recall the standard Datalog semantics.} 

\smallskip\noindent
\change{\textbf{Datalog Semantics}. 
The semantics of Datalog program $P$ can be defined quite easily in terms of model theory~\cite{ceri1989you}. A \textit{Herbrand Base} (HB) is the set of all the facts that we can express in $P$, so all facts of the form $F(c,\ldots,c_k)$, where $c_i$ are constants and $F$ is a predicate. We call EHB, the \textit{Extensional Herbrand Base}, as the subset of HB containing all \changeR{facts} whose predicate $F$ never appears in a rule head of $P$. Conversely, let IHB be the \textit{Intensional Herbrand Base}, so the subset of HB containing all \changeR{facts} whose predicate $F$ appears in a rule head of $P$. Let $S\subseteq P$ be a finite set of Datalog rules. 
\changeR{
Let us discuss what is usually meant by \textit{logical consequence} in the Datalog context. A fact $F$ is a logical consequence of a set of rules $S$ ($S\models F$) if, for every interpretation $I$ that satisfies $S$, it also satisfies $F$. We limit ourselves to the \textit{Herbrand Interpretations} (HI): those that can be built by subsetting the HB. A \textit{Herbrand Model} (HM) for $S$ is then a HI satisfying $S$. Note that, as usual in this context, the closed-world assumption is adopted. }
We name $\textit{cons}(S)$ the set of all facts $F$ that are \textit{logical consequences} of $S$. The semantics of a Datalog program $P$ is a mapping $\mathcal{M}_p$ from \changeR{the powerset of} EHB to \changeR{the powerset of} IHB that associates every possible database $D \subseteq$ EHB to the set $\mathcal{M}_p(D)$ of intensional derived facts defined as $\mathcal{M}_p(D) = \textit{cons}(P \cup D) \cap \text{IHB}$. 
We define $\textit{cons}(S)=\{I~|~I~\text{is a}~\textit{HM}~\text{of}~S\}.$ As the intersection of HMs is a HM, we have that $\textit{cons}(S)$ is the \textit{Least Herbrand Model} (LHM) of $S$.}

\change{With these model-theoretic premises, let us extend our consideration to existential rules, with an operational approach.}
The semantics of a set of existential rules $\Sigma$ over an instance $D$, denoted $\Sigma(D)$, can be defined via the \emph{chase procedure}. The chase adds new facts to $D$ until $\Sigma(D)$ satisfies all the existential rules. The terms of the facts may include freshly generated symbols, namely \textit{labelled nulls} or \textit{marked nulls}, to satisfy existential quantification.

\begin{example}
\label{ex:mother} \textit{Consider the database $D = \{{\rm Person}({\rm Alice})\}$, and the set of existential rules:}
\begin{center}
    $(1)~\textnormal{Person}(x)\to\exists{z}\,\textnormal{HasMother}(x,z).~~~(2)~\textnormal{HasMother}(x,y)\to\textnormal{Person}(y).$
\end{center}
\textit{The ground fact triggers Rule~1, and the following facts are added to $D$ by the chase, where $\nu_1$ is a labelled null:}  $\textnormal{HasMother}({\rm Alice},\nu_1)$ and $\textnormal{Person}(\nu_1)$.
\textit{The fact \textnormal{Person}$(\nu_1)$ triggers again Rule~1, and the chase adds the facts} $\textnormal{HasMother}(\nu_1,\nu_2)$ and $\textnormal{Person}(\nu_2)$.
\textit{where $\nu_2$ is a new labelled null. Finally, the chase result is the following instance, where $\nu_1$,$\nu_2$,... are labelled nulls:}
$\textnormal{Person}({\rm Alice}),\textnormal{HasMother}({\rm Alice},\nu_1)\}\cup_{i>0}\{\textnormal{Person}(\nu_i),\textnormal{HasMother}(\nu_i,\nu_{i+1})$.~~$\blacksquare$
\end{example}

\smallskip\noindent\textbf{The chase}.
Let us see the chase more formally. Initially we have $\Sigma(D) = D$. 
By a \emph{unifier} we mean a mapping from variables to constants or labeled nulls. 
We say $\rho = \varphi(\bar x,\bar y) \rightarrow \exists \bar z\, \psi(\bar x,\bar z)$ is \emph{applicable} to $\Sigma(D)$ if there is a unifier $\theta_\rho$ such that $\varphi(\bar x \theta_\rho,\bar y \theta_\rho)  \subseteq \Sigma(D)$ and  $\theta_\rho$ has not been used to generate new facts in $\Sigma(D)$ via $\rho$. If $\rho$ is applicable to $\Sigma(D)$ with a unifier $\theta_\rho$, then it performs a \emph{chase step}, i.e., it \emph{generates} new facts $\psi(\bar x \theta_\rho',\bar z\theta_\rho')$ that are added to $\Sigma(D)$, where $\bar x \theta_\rho = \bar x \theta'_{\rho}$ and $z_i \theta'_{\rho}$, for each $z_i \in \bar z$, is a  fresh labeled null that does not occur in $\Sigma(D)$.
The chase step easily generalizes to a set of rules.
 The procedure performs chase steps until no rule in $\Sigma$ is applicable. $\Sigma(D)$ is potentially infinite because of the generation of \change{infinitely many} labeled nulls. However, we will consider
the chase up to isomorphism of facts, which is sufficient for 
our logical reasoning task in Warded Datalog$^\pm$
and is finite~\cite{BeSaGo18}, as we shall see.

\smallskip
\noindent
\textbf{Wardedness}. Let us define \change{as} \textit{frontier variables} the universally quantified variables of a rule that also appear in the head. Wardedness introduces syntactic restrictions to limit the propagation of labelled nulls in the frontier variables.
Given a predicate $p$ appearing in a set of rules $\Sigma$, a position $p[i]$ is the $i$-th term of $p$, with $i=1,\ldots$. A position $p[i]$ is \textit{affected} if:
(i)~$p[i]$ contains an existentially quantified variable for a rule $\rho$ of $\Sigma$ (e.g., $\textsf{HasMother}[2]$ in Example~\ref{ex:mother} is affected); or, 
(ii)~for some rule $\rho$ of $\Sigma$ s.t.\ a frontier variable $x$ only appears in affected body positions of $\rho$ and in position $p[i]$ in the head (e.g., $\textsf{Person}[1]$ is affected).

Affectedness induces the following classification of variables. If a variable appears only in affected positions of a rule $\rho$, then it is \textit{harmful}, otherwise it is \textit{harmless}, with respect to that rule. A harmful frontier variable is denoted \textit{dangerous}. For instance, the variable \textit{y} in the second rule of Example~\ref{ex:mother} is dangerous. A rule is \textit{warded} if: (i)~all the dangerous variables appear in the body in a single atom (the \textit{ward}); and, all the \change{variables of the ward} that are in common with other body atoms are harmless.  A program is warded if all its rules are warded. The program in our Example~\ref{ex:mother} is warded and, in particular the wards are $\textsf{Person}$ in the body of the first rule, and $\textsf{HasMother}$ in the body of the second one.

\medskip\noindent
\changeR{\textbf{Warded Reasoning}.}
\change{Let $D$ be a database instance over the domain of constants $\mathbf{C}$, and $\Sigma$ be a program.}
\changeR{Given a \emph{query} $Q = (\Sigma,{\rm Ans})$ where ${\rm Ans}$ is an $n$-ary predicate, an \emph{answer} $Q(D)$ is the set of all facts ${\rm Ans}(\bar t)$, where the tuple ${\bar t} \in (\mathit{dom}(D) \cup \mathbf{N})^n$, such that ${\rm Ans}(\bar t) \in \Sigma(D)$.}

\changeR{First of all, observe that we allow labelled nulls in the answers. In fact, it is often the case that the set of rules $\Sigma$, modeling the domain of interest, cannot completely generate from $D$ all the values of the intensional atoms, and yet we wish to return the labelled null values, for instance for later comparisons between the tuples in the query answer.}

Also, since $\Sigma(D)$ is potentially infinite, the number of answers to a query could be infinite as well. 
\changeR{In this work we choose to find a general representative answer that subsumes all the others. To this end, we borrow the notion of \emph{universal answer}~\changeR{\cite{FKMP05}, very common in data exchange settings.} A universal answer $\mathcal{U}$ is such that for any other answer $\mathcal{U}^\prime$, there exists a substitution of labelled nulls $\theta_U$ such that $\mathcal{U}\theta_U$ (i.e., the application of the substitution to all the facts contained in the answer) coincides with ${U}^\prime$.} Intuitively, \changeR{as it represents the most general answer, a universal answer can be mapped onto any other answer by assigning the labelled nulls appearing in its facts.}
In our setting, a \emph{logical reasoning task} or more simply \textit{reasoning task} amounts to computing a universal answer. For a Warded Datalog$^\pm$ program, the reasoning task is decidable and is \textsf{PTIME} in data complexity (i.e., when the query is fixed and the data size varies)~\cite{GoPi15}.

\change{
Two facts are \textit{isomorphic} when they have the same terms up to renaming of the labelled nulls. More technically, when there exists a bijection between their terms that preserves constants.}
\change{Given a database instance $D$, a set of warded rules $\Sigma$, let $\Sigma(D)/\mathcal{I}$ be the \changeR{quotient structure} induced by the fact isomorphism relation $\mathcal{I}$. We define $\Sigma_{W}(D)$ as \textit{warded semantics} or \textit{warded chase} as the set of all the class representatives of $\Sigma(D)$ in $\Sigma(D)/\mathcal{I}$, one for each equivalence class. \changeR{Roughly speaking, $\Sigma_{W}(D)$ can be seen as a ``flat'' version of a standard quotient structure, where for each equivalence class, a representative is chosen.} We have that $\Sigma_{W}(D)$ is finite, as facts have finite arity and labelled nulls can appear in a finite number of positions, eventually giving rise to isomorphic copies.}

\smallskip\change{Operationally, $\Sigma_W(D)$ is generated by applying} reasoning algorithms that just execute a finite number of chase steps, based on the recognition of ``repeating patterns'' in the chase~\cite{BeSaGo18}. In particular, given two isomorphic facts \(h\) and \(h^\prime\), 
one needs to explore only \(h\) and so never perform chase steps starting from \(h^\prime\). 
\change{In this sense, we will refer to \emph{warded chase step} to mean a chase step limited to those unifiers allowed by the isomorphism criterion.} We will use the term \textit{fact-isomorphic} to refer to database instances for which there exists a bijection between their facts s.t.\ the corresponding facts are isomorphic. \change{The warded chase semantics is uniquely defined for $D$, and independent of the rule application order, modulo fact-isomorphism. In fact: (i) all applicable rules are applied in the chase, and (ii) after a normalization step that eliminates joins on harmful variables, whenever two facts are isomorphic, the chase graph portions derived from them are guaranteed to be isomorphic as well~\cite{BeSaGo18}[Th.2]. In total, no class representatives are lost in any rule application order.}

\smallskip\changeR{
An interesting practical consequence of wardedness is that
queries can be evaluated against the finite instance $\Sigma_W(D)$, with the following guarantee:}
\change{(1)~\textit{a fact $\bar t$ is in $\Sigma_W(D)$ iff $\bar t$ is in $\Sigma(D)$, that is, the usual chase semantics and the warded semantics are equivalent with respect to the reasoning task.} 
\change{The sufficient implication of (1) ($\Rightarrow$) holds since $\Sigma_{W}(D) \subseteq \Sigma(D)$ by construction; the necessary implication of (1) ($\Leftarrow$) directly descends from~\cite[Th.2]{BeSaGo18}: intuitively, the only case in which a fact $\bar t \in \Sigma(D)$ would not correspond to any isomorphic fact $\bar t^\prime \in \Sigma_W(D)$ would postulate the existence of an ancestor $\bar s^\prime\in\Sigma(D)$ for which no warded chase step has been activated in $\Sigma_W(D)$ because of an isomorphic fact $\bar s^{\prime\prime}\in\Sigma_W(D)$ having already triggered a rule. Yet, this would contradict~\cite[Th.2]{BeSaGo18}, which guarantees that the chase subgraph derived from $\bar s^\prime$ and $\bar s^{\prime\prime}$ are isomorphic and so that $\bar t^\prime \in \Sigma_W(D)$.}}

\changeR{It is also worth remarking that, besides atomic  queries (only one body predicate), a larger class of conjunctive queries, namely \textit{warded conjunctive queries}, can be evaluated against the finite instance $\Sigma_W(D)$, with a query equivalence guarantee that directly descends from~\cite[Th.2]{BeSaGo18}. 
We say that a conjunctive query $Q : $ $q(\mathbf{x}) \leftarrow \boldsymbol{\phi}(\mathbf{x},\mathbf{y})$, where $\boldsymbol{\phi}(\mathbf{x},\mathbf{y})$ is a conjunction of atoms, and $\mathbf{x},\mathbf{y}$ are tuples of terms in $\mathit{dom}(D) \cup \mathbf{V}$, is warded w.r.t. a set of rules $\Sigma$, if $\Sigma \cup \{\rho_Q\}$ is warded, where ${\rho_Q}$ is a rule expressing $Q$ as $\nu(\mathbf{x}) \leftarrow \boldsymbol{\phi}(\mathbf{x},\mathbf{y})$ and $\nu$ is an invented atom. Operationally, if not already atomic, the conjunctive query $Q$ is translated into a rule $\rho_Q$ of $\Sigma$ and simulated by the atomic query $\hat{Q}:q(\mathbf{x})\leftarrow\nu(\mathbf{x})$.}

\smallskip\noindent
\change{\textbf{Additional Features}.} \vadalog is a logical language based on Warded Datalog$^\pm$ and extending it with features of practical utility such as negation, aggregation, algebraic operators, and so on. The language is fully implemented and engineered in the \vadalog system~\cite{BeSaGo18}, a KG management system. \change{While in this work we concentrate on the core of \vadalog, Warded Datalog$^\pm$, we will use negation and aggregations in the application cases.}

\change{\vadalog adopts the usual \textit{stratified negation} semantics~\cite{DEGV01}, the principle of which is operationally straightforward: provided that a negated atom ``$\textit{not}~a$'' appearing in a rule $\rho$ of $\Sigma$ is \textit{safe}, i.e., has as terms only variables appearing in \changeR{positive} body atoms of $\rho$, for each unifier for which, not considering $a$, $\rho$ is applicable, the rule body is evaluated only if such unifier does not bind $a$ to any fact that has already been generated.}
\changeR{Intuitively, the stratification guarantees that negation is not used in a circular way in the dependency predicate graph of $\Sigma$. Hence, in the evaluation of ``$\textit{not}~a$'', the extension of $a$ can be entirely and unambiguously determined beforehand. The stratification condition can be syntactically checked.}

Aggregations have been introduced in multiple contexts using a logical formalism and a careful definition of their semantics arose in all of them, especially when procedural semantics is used for operational reasons besides a model-based one.
Here, we adopt a simple solution, based on stratified semantics~\cite{MuPR90}, which is however enough for our purposes. For each unifier for which a rule is applicable, the aggregation condition is evaluated on all the possible unifiers (defining its operands) that bind the rule body to facts that have already been generated. \change{A rule $\varphi(\bar{x},\bar{y}), v=\textit{aggr}(q) \to \psi(\bar{x},v)$, where $q$ is a variable of $\bar{x}$ and \textit{aggr} is a generic aggregation operator, the value of variable $v$ is computed by aggregating the values of $q$ over the distinct groups defined by the values of \changeR{$\bar{x}\setminus\{q\}$.}}

\smallskip
\section{A Framework for Probabilistic Reasoning with Knowledge Graphs}
\label{sec:pkgs}

In this section, we are going to introduce Probabilistic Knowledge Graphs (PKGs) and \svadalog. Let us outline the approach first.      
A rule $\rho :  \forall\bar x \forall\bar y (\varphi(\bar x,\bar y)\ \rightarrow\  \exists\bar z \, \psi(\bar x, \bar z))$ of a program $\Sigma$ can be viewed as a hard constraint over~$D$. \change{The chase of $D$ under $\Sigma$ enforces all the constraints in $\Sigma$, by applying rules until they are all satisfied. Query answering is then performed on the database instance derived in this way.}

Probabilistic Knowledge Graphs soften this constraint and admit incomplete answers derived from a partial chase. A partial chase can be seen as a derived database not satisfying some of the rules of $\Sigma$ or, in operational terms, the chase steps needed to generate the conclusion of some rules of $\Sigma$ have not been applied. Along the lines of Markov Logic Networks, in PKGs, each rule has an associated weight \change{representing the difference in log probability between a database instance that satisfies the rule vs.\ one not satisfying it. In this sense, the rule weights are not probabilities, but measure the importance of a rule in $\Sigma$.}

More precisely, PKGs induce a probability distribution on the facts in all the partial chases $\Sigma(D)$, such that their likelihood is proportional to the weight of the rules that have been applied.

\smallskip\noindent\textbf{Approach Outline}.
Our approach consists of the following two pillars: (i) we define a Probabilistic Knowledge Graph as the \change{combination} of an input database $D$ and a set \(\Sigma\) of probabilistic existential rules expressed in \svadalog (a probabilistic extension of \vadalog), (ii) 
we construct a structure, called \emph{chase network}, that comprises all possible databases that can be obtained from $D$ by applying rules in $\Sigma$. This structure is already enough to compute marginal probabilities of the facts in the answer of some given query $Q$ and solve the probabilistic reasoning tasks. Points (i) and (ii) are dealt with in Section~\ref{sec:pkgs_and_beyond}.

However, in order to make the framework applicable, we need to mitigate two issues: logical inference in the presence of general FO rules is undecidable or intractable; computing exact marginal probabilities is intractable as well (\textsf{\#P-hard}). For the first issue, we leverage Warded Datalog$^\pm$, for which reasoning is polynomial, as we have seen. For the second, we compute \emph{approximate} marginal probability. 
Thus, in Section~\ref{sec:mcmc-chase} we introduce an MCMC method that simultaneously performs logical and marginal inference. 

\subsection{Soft Vadalog and Probabilistic Knowledge Graphs}
\label{sec:pkgs_and_beyond}

\change{We extend Warded Datalog$^\pm$, and as a consequence \vadalog,} to \svadalog by introducing soft rules. A \svadalog \textit{rule} is a pair $(\rho, w(\rho))$, where $\rho$ is a warded rule and $w(\rho)\in \mathbb{R} \cup \{+ \infty, -\infty\}$ is a \emph{weight}, reflecting how strong a constraint is and so the absolute bias for a model to respect it (or not to respect it, in the case of negative weights).
A soft rule $(\rho, + \infty)$  is called a \emph{hard} rule. 
A \svadalog~\emph{program} is a set of \svadalog~rules.

\medskip\noindent\textbf{Semantics of \svadalog}. A \svadalog program specifies
a probability distribution over a \textit{chase network}, a graph holding the database instances generated by all the partial chase applications over a given database instance.

\smallskip
\emph{A \emph{Probabilistic Knowledge Graph} is a pair $\langle D, \Sigma \rangle$, where $D$ is a database instance \change{over the domain of constants $\mathbf{C}$} and $\Sigma$ is a \svadalog~program.}
A PKG can be viewed as a template for constructing  \emph{chase networks}. 
We define the \textit{closure} $\mathit{cl}_\Sigma(D)$ of a database $D$ under a set of warded rules $\Sigma$ as the database $D^\prime$ obtained by computing the chase \change{$D^\prime = \Sigma^\prime_W(D)$, where $\Sigma^\prime \subseteq \Sigma$ is the set of the hard rules in $\Sigma$.} By wardedness, the database $D^\prime$ is finite and unique modulo-fact isomorphism \change{(Section~\ref{sec:kgs_action}).}
$D$ is \textit{closed under $\Sigma$} if $D$ is fact-isomorphic to \change{$\Sigma_W(D)$,} i.e., \change{$D = \Sigma_W(D)$.}
Given a PKG $\mathcal{G} = \langle D,\Sigma\rangle$, a \emph{chase network} $\Gamma(\mathcal{G})$ 
\changeR{is a triple  $\langle \mathbf{W}, \mathbf{T}, W_0  \rangle$, where:}
\begin{enumerate}
   \item $\mathbf{W}$ is a set of nodes, $\mathbf{T}$ is a set of edges. 
   \item \changeR{Each node of $\mathbf{W}$ corresponds to a class $C$ of all the \textit{reachable databases instances}, each closed under the hard rules of $\Sigma$ and with relation symbols from $D\cup\Sigma$. In particular, each class $C$ represents a set of fact-isomorphic database instances.}
   
   \item \changeR{ \emph{$W_0 \in \mathbf{W}$ is a source node associated to $cl_\Sigma(D)$, i.e., $W_0$ is associated to the closure of $D$.}}
    \item \emph{There is an edge $t_s \in \mathbf{T}$ 
    from $W$ to $W'$ iff the database instance associated to
    $W'$ can be obtained from the one associated to $W$ 
    by one transition step. 
    A \emph{transition step} $s$ from 
    $W$ to $W'$ consists of a warded chase step of at least 
    one applicable soft rule with one unifier followed by the closure with respect to the hard rules of $\Sigma$.
    Edge $t_s$ is then labeled by $\sum_{\rho\in \sigma} w(\rho)$, where $\sigma$ is the set of soft rules applied.}
\end{enumerate}

\noindent

\noindent A node $W$ of $\Gamma$ represents a class of fact-isomorphic databases. 
\changeR{By wardedness, 
$\mathbf{W}$ and each database associated to a node $W\in\mathbf{W}$ are \textit{finite}.}  
Each class of fact-isomorphic instances is represented by exactly one node, and all the paths of the chase network connecting $W_0$ to the same class of instances converge into the same terminal node. In fact, multiple isomorphic versions of the same instances should not be considered at all because their facts are semantically equivalent for query answering and so should be for marginal probability. Nevertheless, for its computation, we will take into account the influence of all possible transition steps leading to isomorphic instances.
Each transition step can only add facts to a node $W_i$. Therefore the chase network does not contain directed cycles.
Moreover, the chase network is a \textit{multigraph}, since two nodes $W$ and $W^\prime$ can be connected by multiple edges, one for each possible transition step from $W$ to $W^\prime$.

\smallskip
\change{Let $\Pi_\mathcal{G}(W)$ be the set of all the edges appearing in any path from $W_0$ to a node $W$ in $\Gamma(\mathcal{G})$.
We define the \emph{weight} $w(W)$ of $W$ 
as the sum of the edge labels of the edges in $\Pi_\mathcal{G}(W)$.} 
The chase network induces the following probability distribution over its nodes: 

\begin{equation}\label{eq:weight}
P(W) = \frac{1}{Z}~{\rm exp}~w(W).
\end{equation}

\smallskip\noindent
\change{By considering the summation of the weights along all the possible paths to a node $W$, we want to capture the relevance of $W$ with respect to the entire derivation process, so not only which rules/unifiers it satisfies, but also how ``reachable'' it is in a chase execution.} 
The normalization constant $Z$ is a \emph{partition function} to make $P(W)$ a proper distribution. It is defined as \(Z= \sum_{W} ~{\rm exp}~w(W)\). 
For a given fact \(f\), its \emph{marginal probability} $P(f)$ can be calculated
as the summation of the likelihood of the nodes it appears in 

\begin{equation}\label{eq:marginal_probability}
P(f) = \sum_{W_i : f\in W_i} P(W_i).
\end{equation}

\smallskip
\noindent
Figure~1 shows a fragment of a possible chase execution (left part) and the respective chase network (on the right) for \change{the inductive definition in} Example~\ref{ex:second}. In the chase execution diagram, the nodes are the facts $f$ in the database instances associated to nodes $W_i$ of the chase network. Facts $f$ are annotated with a set $\{W_0,\ldots,W_n\}$ of nodes of the chase network such that for each $W_i$ in the set, $f \in W_i$. Solid edges are warded chase steps applying hard rules; dashed edges are for soft rules, with weight $\gamma$. In the chase network, nodes are database instances; they are connected by edges annotated with the rule $(\rho)$ that has been applied and its weight (in boldface). 

\medskip


\change{\begin{figure}
\begin{center}
\hspace{5mm}\includegraphics[scale=0.35]{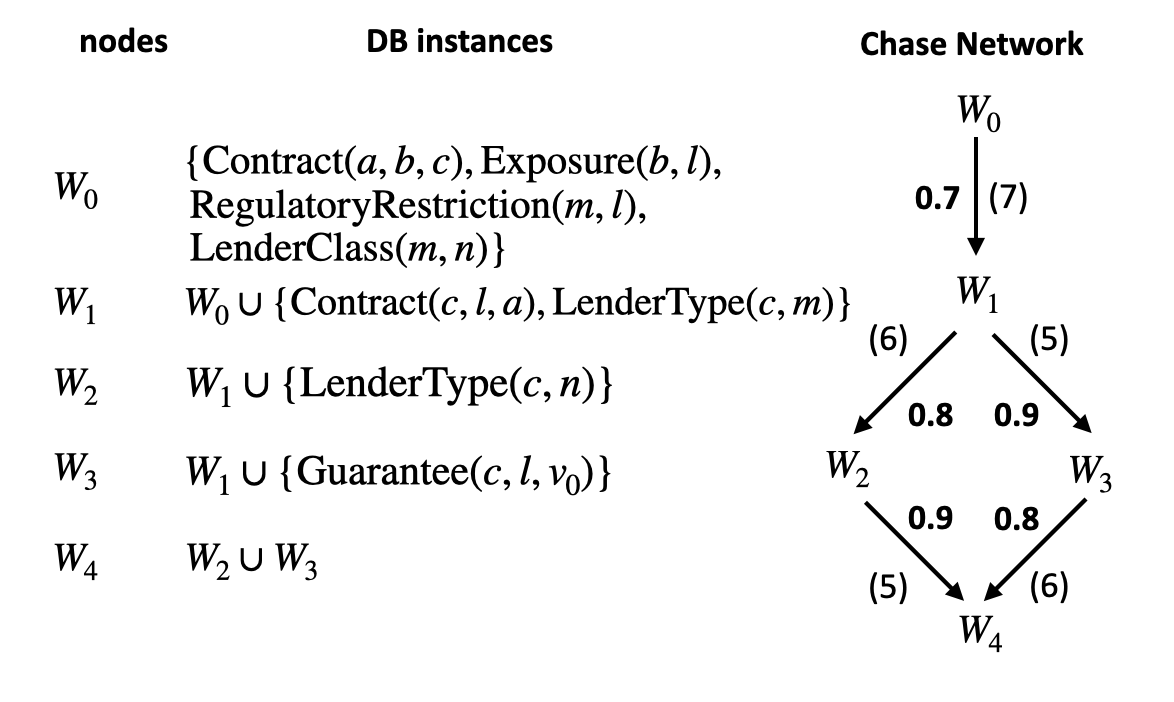} 
    \label{fig:chase_network}
\caption{Chase execution and chase network for Example~\ref{ex:first}.}
\end{center}
\vspace{-2mm}
\end{figure}}
From the chase network, we can now compute the marginal probability of \textsf{Contract}s.
By applying Equation~\ref{eq:weight}, we have $w(W_0)=0$. Then it follows
$w(W_1)=0.7$ and $w(W_2)=0.7+0.8=1.5$, $w(W_3)=0.7+0.9=1.6$, and $w(W_4)=0.7+(0.8+0.9)\times 2=4.1$.
So we can calculate marginal probability for \textsf{Contract}$(c,l,a)$ with Equation~\ref{eq:marginal_probability}. This fact appears 
for $W_1$, $W_2$, $W_3$, $W_4$, so we have: $(e^{0.7}+e^{1.5}+e^{1.6}+e^{4.1}) / Z = 0.99$, with $Z=1+e^{0.7}+e^{1.5}+e^{1.6}+e^{4.1}$.
Similarly, for \textsf{Contract}$(c,l,v_0)$, we have: $(e^{1.6} + e^{4.1}) / Z = 0.9 $.

\smallskip\noindent\change{\textbf{Discussion}. As our framework is based on Datalog$^\pm$ features, it incorporates existential quantification with the expressive power allowed by the warded fragment (see Section~\ref{sec:kgs}), when also recursion is involved, and so suitable for ontological reasoning (see requirement~(i) of Section~\ref{sec:introduction}).
We use the chase as an operational tool to generate all the facts induced by a database instance and a Warded Datalog$^\pm$ program. \changeR{Unlike in MLNs, the program can encode inductive definitions,} e.g., involving transitive closures, and supports full recursion, as directly inherited from Datalog semantics. In fact, according to the model-theoretic semantics of Datalog, the generated facts are those defining the Least Herbrand Model of the program (see Section~\ref{sec:kgs}), and the facts that, while satisfying the program rules, are not directly entailed by the chase---and so not part of the LHM---are not derived. Unlike in MLNs, these facts therefore assume zero probability by construction (requirement~(ii)).}

\smallskip
\change{It is worth making a final note about the origin of \svadalog programs. While there is substantial soaring literature about first-order rule learning, we consider it out of scope here and assume that rules are defined by domain experts. As weights reflect the relative importance of rules in the KG, in our experience, they can be effectively assigned and tuned by domain experts as well, within an iterative process. They can also be efficiently learnt from relational data by optimizing a pseudo-likelihood measure: while they cannot be estimated with maximum-likelihood, because of the concavity of our log-likelihood function, they can derive from gradient-based, quasi-Newton estimations or iterative scaling~\cite{NoWr99}, like in MLNs.}

\subsection{Probabilistic Reasoning}
\label{sec:prob_reasoning}

Let us now formally study the probabilistic reasoning task. 
\change{Given a probabilistic knowledge graph $\mathcal{G} =\langle D,\Sigma\rangle$, where ${\rm Ans}(\bar t)$ is an $n$-ary predicate, let \(Q = (\Sigma,{\rm Ans})\) be a query.} \changeR{A pair $\langle {\bar t}, p \rangle$, where ${\bar t}$ is a tuple and $p \in [0,1]$ is a real number, is in a \textit{probabilistic answer} $Q(D)$ if ${\rm Ans}(\bar t)$ is a fact of some instance associated to nodes of $\mathbf{W}$ in $\Gamma(\mathcal{G})$, and $p$ is the marginal probability $P(\bar t)$ of ${\rm Ans}(\bar t)$. Then, the probabilistic reasoning task consists in computing the probabilistic answer $Q(D)$ as the set of facts $\{ \langle {\bar t}, P(\bar t) \rangle \}$.}

\smallskip\noindent Probabilistic reasoning consists of two phases: (i) \textit{grounding}, that is, the construction of the chase network; (ii) \textit{marginal inference}, that is, the computation of the marginal probability for each fact in the query answer. 
Grounding requires an exponential number of chase executions,  each with polynomial complexity---by wardedness---in the size of $D$. 
Marginal inference in \svadalog is \textsf{\#P-hard} when the program is assumed to be fixed. 

 \begin{proposition}
 \textit{The probabilistic reasoning task is $\#P$-hard in data complexity.}
 \end{proposition}
\begin{proof}
We adapt the proof for $\#P$-hardness of query answering over probabilistic databases from~\cite{SORK11}.
Let $\Phi = \bigvee_{(i,j)\in E} X_i \wedge Y_j$ be a PP2DNF formula, where $X_1, X_2, \ldots,$ and $Y_1, Y_2\ldots$ are disjoint sets of propositional variables. It is known that the problem of counting the number of satisfying assignments for PP2DNF formulas is hard for $\#P$\change{~\cite{PrBa83}.}
Suppose we have three EDBs $R/1, T/1$ and $S/2$. 
We define EDB $D$ as the set of facts $\{R(X_1), R(X_2),\ldots,$  $T(Y_1), T(Y_2),\ldots, \} \cup \{S(X_i, X_j) \mid (i, j) \in E\}$.
Additionally we define a set of rules:
\begin{align*}
&0 :: R(x) \rightarrow R'(x), \\
&0 :: T(x) \rightarrow T'(x), \\
&R'(x), S(x,y), T'(y) \rightarrow Q().
\end{align*}
\change{Intuitively speaking, weight $0$ of both rules ensures that all possible worlds have equal weights. With worlds having equal weights, computing probabilities comes down to counting, as is our goal. Let us now give the details.}
We associate an assignment $\theta$ for variables $X_i$ and $Y_j$ with a possible world $\mathcal{D}$ as follows:
$\theta(X_i) = \mathit{true}$ iff $R'(X_i) \in \mathcal{D}$ and $\theta(Y_j) = \mathit{true}$ iff $T'(Y_j)  \in \mathcal{D}$, which establishes 1-1 correspondence.
Note that $Q() \in \mathcal{D}$ iff $\Phi$ is evaluated to true under $\theta$.
Then the number of satisfying assignments $q$ for $\Phi$ equals $2^n \cdot P(Q())$, where $n$ is the number of the Boolean variables $X_i$ and $Y_j$.
\change{In fact, $P(Q()) = q\cdot e^0 / Z = q/2^n$, and $Z=2^n\cdot e^0$.}
Thus an algorithm that computes marginal probability $P(Q())$ also computes the number of satisfying assignments for $\Phi$ which is $\#P$-hard.
\end{proof}

\section{The MCMC-chase Algorithm}
\label{sec:mcmc-chase}

In order to cope with the high complexity of reasoning in PKGs, we introduce the MCMC-chase, a technique that blurs the conceptual distinction between grounding and marginal inference and performs marginal inference by sampling the chase network. Instead of performing the full grounding of the chase network and then sampling it, we make the chase driven by an MCMC algorithm so that only a representative ``subspace'' of the chase network---a subgraph---is built. On that subgraph, the MCMC-chase computes exact marginal inference by applying Equation~\ref{eq:marginal_probability}. The relevant subgraph of the chase network is chosen in dependence of the weights of the rules and, as a consequence, of the induced instances for the nodes in $\mathbf{W}$. 

\medskip
In particular, the MCMC-chase is an independence sampling\change{~\cite{Tie94}} MCMC where the chase procedure is seen as a \textit{Markov process}~\change{\cite{gilk95}} over the nodes of the chase network.
Given a PKG $\mathcal{G} = \langle D,\Sigma\rangle$, the MCMC-chase  
applies soft rules of $\Sigma$, with a probability that is proportional to the rule weight and generates nodes of $\Gamma(\mathcal{G})$. 
The algorithm starts from $D$ and applies rules, creating new nodes of $\mathbf{W}$. The algorithm keeps track of the weight of the current node and decides to whether accept or reject it according to an \emph{acceptance probability}, in a Metropolis-Hastings\change{~\cite{hast70}} style. After a fixed number $N$ of iterations, the algorithm stops and returns the probability for all the generated instances computed by Equation~\ref{eq:weight}, so that Equation~\ref{eq:marginal_probability} can be applied to determine the marginal probability of the facts. Higher values for $N$ result into deeper chase networks, and thus more precise marginal probability. If $N$ is high enough to compute the full chase, the precision is maximum as MCMC-chase degenerates into exact marginal inference over the chase network.

\algrenewcommand\alglinenumber[1]{\tiny #1:}
\algnewcommand{\IfThenElse}[3]{
  \State \textbf{if} #1\ \textbf{then} #2 \algorithmicelse\ #3}

\begin{algorithm}[]
\caption{MCMC-chase  \label{alg:mcmc_chase}}
 \begin{algorithmic}[1]
     \Statex
      \Function{mcmc-chase}{$\mathcal{G} = \langle D,\Sigma \rangle$,$N$}
                  \State \(\mathbf{W}_S  = \emptyset \)
                  \Comment samples from the distribution over the nodes $\mathbf{W}$ of $\Gamma(\mathcal{G})$
            \State \(\mathcal{D}^0  = D \)
  
      \For {$n \leftarrow 1$ to $N$ }
      \Comment{$N$: \# of iterations}

      	\State Sample \(S  \sim~\mathcal{P}(\lambda) \)
      	\Comment{\# of steps, from a Poisson distr.}
		\State \(\mathcal{T} \leftarrow \mathcal{D}^{n-1}\)	
		\State \(w(\mathcal{T}) \leftarrow w(\mathcal{D}^{n-1})\)	

		\For {$s \leftarrow 1$ to $S$ }
	 		 \State Sample \(\delta \sim~\mathcal{U}(0,1)\); Sample \(\mu \sim~\mathcal{U}(0,1) \)

			\State \(\mathbf{R_a} \leftarrow\) all applicable  $\rho$ in \(\Sigma\) s.t. \(\mu < 1-e^{-w(\rho)} \), \change{with randomly chosen unifier $\theta_\rho$}
			\State \(\mathbf{R_u} \leftarrow\) all undoable  $\rho$ in \(\Sigma\) s.t. \(\mu < 1-e^{-w(\rho)} \), \change{with randomly chosen unifier $\theta_\rho$}

        			\IfThenElse{$\delta < 0.5$}
        			{{\sc transition\_step}\((\mathcal{T},\mathbf{R_a})\)~~~~~~~~~~~~~~~~~\\~~~~~~~~~~~~~~~~~~~~~~~~~~}
        			{{\sc undo\_transition\_step}\((\mathcal{T}, \mathbf{R_u})\)}



		\EndFor      	

      		\State \(\alpha \leftarrow {f(\mathcal{T})}/{f(\mathcal{D}^{n-1})}\)
      		\Comment{acceptance probability}
      		
		\State With prob. $\min(1, \alpha)$, accept and add \(\langle \mathcal{T},w(\mathcal{T})\rangle\) to \(\mathcal{\mathbf{W}_S}\) 
		\IfThenElse{accepted}{ \(\mathcal{D}^{n} \leftarrow \mathcal{T}\)}{ \(\mathcal{D}^n \leftarrow \mathcal{D}^{n-1}\)}
     			\Comment{accept or rollback}
     \EndFor

     \Return \(\mathbf{W}_S\)
      
      \EndFunction
  \end{algorithmic}
\label{algo:mcmc}
\normalsize
\end{algorithm}

\noindent
Algorithm~\ref{algo:mcmc} gives pseudo-code for the MCMC-chase.
It takes as input a PKG $\mathcal{G}$ and returns
samples from the distribution $P(W) = \frac{1}{Z}~{\rm exp}~w(W)$
over the nodes $\mathbf{W}$ of the chase network $\Gamma(\mathcal{G})$.
The algorithm performs $N$ iterations, each consisting of $S$ steps,
with $S$ extracted from a Poisson (\emph{jump}) distribution (line 5).
In each step, forward or backward depending on a value \(\delta\) uniformly
chosen, the algorithm selects subsets
$\mathbf{R_a}$ and $\mathbf{R_u}$ of rules from \(\Sigma\) with a probability proportional
to $w(\rho)$ (lines 10-11) of \textit{applicable} or \textit{undoable} rules. \changeR{This is obtained by uniformly choosing $\mu$ in the $(0,1)$ interval and checking whether $\mu<1-e^{-w(\rho)}$, whose likelihood of being satisfied grows proportionally with the weight of $\rho$, as the amount $e^{-w(\rho)}$ decreases.} \change{For simplicity, we are only considering positive weights in the pseudocode as the extension to negative ones is straightforward (i.e., $\mu > 1-e^{w(\rho)}$).}
As defined in detail in Section~\ref{sec:kgs}, a rule $\rho = \varphi(\bar x,\bar y) \rightarrow \exists \bar z\, \psi(\bar x,\bar z)$ is applicable if for some unifier $\theta_\rho$, a warded chase step can produce new facts not in $\Sigma(D)$ via $\rho$. Vice versa, it is undoable if (i) there is a fact $\psi(\bar x \theta_\rho',\bar z\theta_\rho') \in \Sigma(D)$ generated by $\rho$ with some body unifier $\theta_\rho$ extending to $\theta_\rho^\prime$ as $\bar x \theta_\rho = \bar x \theta'_{\rho}$ and $z_i \theta'_{\rho}$ for each $z_i \in \bar z$ being a fresh labeled null of $\Sigma(D)$, and, (ii) there are no facts $\psi^\prime(\bar x \delta_{\rho^\prime}',\bar z\delta_{\rho^\prime}') \in \Sigma(D)$ generated by some rule $\rho^\prime = \psi(\bar x,\bar y) \rightarrow \exists \bar z\, \psi^\prime(\bar x,\bar z)$ with body unifier $\delta_{\rho^\prime}$ extending to $\delta^\prime_{\rho^\prime}$ as $\bar x \delta_{\rho^\prime} = \bar x \delta'_{\rho^\prime}$ and $z_i \delta'_{\rho^\prime}$ for each $z_i \in \bar z$ being a labeled null of $\Sigma(D)$. Intuitively, a rule is undoable if it has not been used by any rule to generate new facts for $\Sigma(D)$ and thus it is a leaf of the chase network.

\emph{Forward transition steps} (line 12) try to apply a transition step with the selected applicable rules $\mathbf{R_a}$ to the current node \(\mathcal{T}\) of the chase network.
\emph{Backward transition steps} (line 13) try to undo a transition step with the selected undoable rules in $\mathbf{R_u}$. Algorithm~\ref{algo:transition} gives the pseudocode for both. In the forward case, for each selected soft rule $\rho$, a unifier $\theta$ is uniformly chosen from the existing ones, and a warded chase step applied; this results in an updated instance $\mathcal{T}$ (lines 2-3). Instance  $\mathcal{T}$ is then updated with its closure with respect to the hard rules in $\mathbf{R}$ (line 4) and its weight incremented by the weights of all the applied soft rules. On the other hand, in the backward case, first the facts generated by the hard rules of $\mathbf{R}$ are removed, with a process that is intuitively a \textit{backward closure}, then, the effects of all the soft rules of $\mathbf{R}$ are canceled, in the sense the facts they generated are removed from $\mathcal{T}$ and their weights subtracted accordingly. Note that hard rules never affect the total weight.
After $S$ steps, an \emph{acceptance function} $f(\mathbf{Y}) = {\rm exp}~w(\mathbf{Y})$ evaluates the acceptability of the current node (lines~14-15 of Algorithm~\ref{alg:mcmc_chase}) in a Metropolis-Hastings style.
Finally, all accepted nodes and their weights are returned.

\begin{algorithm}[t]
\caption{Transition and Undo Transition}
 \begin{algorithmic}[1]
     \Statex
      \Procedure{{\sc transition\_step}}{$\mathcal{T},\mathbf{R}$}
       \For {\change{each soft $\rho \in \mathbf{R}$, use unifier $\theta_\rho$ for $\rho$ }}
	  \State \(\mathcal{T} \leftarrow \mathit{warded\_chase\_step}(\mathcal{T}, \mathbf{R})\) with \change{$\theta_\rho$}
      	  \State $\mathcal{T} \leftarrow cl_{\mathbf{R}}(\mathcal{T})$

     \State $w(\mathcal{T}) \leftarrow w(\mathcal{T}) + \sum_{\rho \in \mathbf{R}}w(\rho)$      
      \EndFor\EndProcedure
      
       \Procedure{{\sc undo\_transition\_step} }{$\mathcal{T},\mathbf{R}$}
       \State Remove facts generated by hard rules of $\mathbf{R}$ from $\mathcal{T}$
       \For {each soft $\rho \in \mathbf{R}$ \change{applied with unifier $\theta$}}

\State Remove facts generated by $\rho$ \change{with $\theta_\rho$} from $\mathcal{T}$      
   \State $w(\mathcal{T}) \leftarrow w(\mathcal{T}) -  w(\rho_i)$

\EndFor
      
      \EndProcedure    
  \end{algorithmic}
\label{algo:transition}
\normalsize
\end{algorithm}

\medskip

Observe that 
the stochastic process underlying the MCMC-chase is a \textit{Markov process} or, equivalently, that it satisfies the \textit{Markov property}. In fact, the MCMC-chase is \textit{memoryless}, in the sense that a future process status only depends on the present one: a candidate node inherits all the facts only from one previously generated node and some facts are added to or removed from it by the applicable (resp.\ undoable) rules. The Markov process associated to the MCMC-chase also has favourable properties, namely \textit{detailed balance} and \textit{ergodicity}. 
\changeR{A Markov process has detailed balance if the transition probabilities respect the following law: $P(\mathcal{D}^i)P(\mathcal{D}^i \rightarrow \mathcal{D}^{i+1})$ between each pair of states $\mathcal{D}^i$ and $\mathcal{D}^{i+1}$ is equal to the transition probability $P(\mathcal{D}^{i+1})P(\mathcal{D}^{i+1} \rightarrow \mathcal{D}^{i})$, where $P(\mathcal{D}^{i})$ and $P(\mathcal{D}^{i+1})$ are the equilibrium probabilities of being in the states $\mathcal{D}$ and $\mathcal{D}^{i+1}$, respectively~\cite{Stuart1991}.}
Intuitively, detailed balance guarantees that the probability of flowing from one node $W_i$ to a connected node $W_j$ of the chase network via applying a forward transition step is equivalent to the probability of flowing from $W_j$ back to $W_i$ by applying a backward transition step. Ergodicity ensures the absence of blocking ``trap states'' so that all the nodes of the chase network are eventually visited.

\begin{proposition}
\textit{The Markov chain generated by MCMC-chase satisfies detailed balance and ergodicity.}

\begin{proof}
First we prove the property of detailed balance.
Let $\mathcal{D}$ be a state at an iteration of the for-loop in lines 8-17 of Algorithm~\ref{alg:mcmc_chase}, and $\mathcal{D}'$ is a state that is the result of either applying or undoing rules $R$ in line 12 or 13. \change{With $P(\mathcal{D}^i)$ being the probability of state $\mathcal{D}^i$,} we then denote the probability of selecting the rules $R$ to be applied or undone to go from $\mathcal{D}$ to $\mathcal{D}'$ as
$P'(\mathcal{D} \rightarrow \mathcal{D}')$ which is also equal to $P'(\mathcal{D}' \rightarrow \mathcal{D})$. 
The next sample state \(\mathcal{D}^{i+1}\) is reachable from the current state \(\mathcal{D}^i\) by $S$ steps of applying or undoing warded chase steps. 
The transition probability $P(\mathcal{D}^i \rightarrow \mathcal{D}^{i+1})$ can be written as
$
\min(1, \alpha) \cdot L(\mathcal{D}^i \rightarrow \mathcal{D}^{i+1}),
$
where $\alpha = \frac{ f(\mathcal{D}^{i+1})}{ f(\mathcal{D}^i)}$ and $L(\mathcal{D}^i \rightarrow \mathcal{D}^{i+1})$ denotes 
$
\sum_{\mathcal{D}^i = D_1, \ldots, D_S = \mathcal{D}^{i+1}} \prod^{S-1}_{j = 1} P'(D_j \rightarrow D_{j+1}).
$

Here the sum is over all possible paths in the chase network from $\mathcal{D}^i $ to  $\mathcal{D}^{i+1}$ of length $S$.
Note that $L(\mathcal{D}^i \rightarrow \mathcal{D}^{i+1})  = L(\mathcal{D}^{i+1} \rightarrow \mathcal{D}^{i})$ since paths between the states are undirected. 
Then\\

$P(\mathcal{D}^i) \cdot P(\mathcal{D}^i \rightarrow \mathcal{D}^{i+1}) = $\\
$\frac{1}{Z} f(\mathcal{D}^i) \cdot  \min(1, \frac{ f(\mathcal{D}^{i+1})}{ f(\mathcal{D}^i)}) \cdot L(\mathcal{D}^i \rightarrow \mathcal{D}^{i+1}) = $\\
$\frac{1}{Z} f(\mathcal{D}^{i+1}) \cdot  \min(1, \frac{f(\mathcal{D}^{i})}{ f(\mathcal{D}^{i+1})})\cdot L(\mathcal{D}^i \rightarrow \mathcal{D}^{i+1})  = $\\
$\frac{1}{Z} f(\mathcal{D}^{i+1}) \cdot  \min(1, \frac{f (\mathcal{D}^{i})}{ f(\mathcal{D}^{i+1})})\cdot L(\mathcal{D}^{i+1} \rightarrow \mathcal{D}^{i}) = $\\
$P(\mathcal{D}^{i+1}) \cdot P(\mathcal{D}^{i+1} \rightarrow \mathcal{D}^{i}).$\\[2mm]

\noindent
Let us show ergodicity. For this it is enough to show that it is possible to reach any state from any other state with non-zero probability.
For any two possible nodes $\mathcal{D}$ and $\mathcal{D}'$ there is a path in the chase network. This path represents applying or undoing of warded chase steps. Let $S'$ be the length of such a path. There is a non-zero probability that the number $S'$ is sampled in line 5 and that exactly the same chase applications or undoing are performed in lines 12 and 13, and that finally the state is accepted in line 16. Therefore the MCMC-chase satisfies ergodicity.
\end{proof}
\end{proposition}

\section{Application Use Cases}
\label{sec:use-cases}

\textit{Record linkage} and \textit{data fusion} are two relevant faces of information integration, both concerned with heterogeneity at instance level. 
Probabilistic knowledge graphs offer a well-founded and integrated framework for such problems.
In this section, we discuss two relevant use cases. 

\subsection{Record Linkage}
\label{sec:record-linkage} 

Record linkage consists in deciding which records of a database refer to the same real-world entities. It plays a crucial role in the standard information integration~\cite{Ullm97}, data mining~\cite{KDD03} and numerous industrial applications~\cite{Chri12}. 
Beyond the seminal statistical approaches~\cite{FeSu69,AgKa11}, more modern techniques~\cite{SiDo05,CuMc05} aid the decision about the matching of one specific pair of entities with decisions about other pairs, even with transitive closure~\cite{McWe04}. MLN frameworks for record linkage effectively generalize the mentioned techniques~\cite{SiDo06}, but inherit the semantic limitations discussed in Section~\ref{sec:relwork}. We show how our framework can handle this domain.

\begin{example}\textit{Consider a Knowledge Graph $G$ with facts describing a network of companies to be matched. It has the following predicates: $\textsf{Company}({\rm company})$, $\textsf{ Industry}({\rm company},{\rm industry})$, $\textsf{Size}({\rm com}\-{\rm pany},{\rm size})$ where the size is in terms of known number of employees (e.g., 1-10, 11-50 employees, etc.), $\textsf{Group}({\rm company},{\rm group})$, denoting the participation of a company in a group, $\textsf{Subsidiary}({\rm company},{\rm company})$, if the first company is a subsidiary of the second, $\textsf{SameSize}({\rm si}\-{\rm ze},$ ${\rm size})$, representing approximate equivalence of company dimensions. Finally, $\textsf{Match}({\rm com}\-{\rm pany},$ ${\rm com}\-{\rm pany})$ witnesses two identical companies. We extend $G$ with the following rules:}
\begin{align*}
0.5 :: {\rm Company}(x),{\rm Industry}(x,z),{\rm Company}(y),{\rm Industry}(y,z) \rightarrow {\rm Match}(x,y) & ~~(1)\\
0.3 :: {\rm Company}(x),{\rm Size}(x,z),{\rm Company}(y),{\rm Size}(y,w),{\rm SameSize}(z,w) \rightarrow {\rm Match}(x,y) & ~~(2)\\
0.9 :: {\rm Company}(x),{\rm Company}(y),{\rm Size}(x,z),{\rm Size}(y,w),\mid x-w \mid<\epsilon \rightarrow {\rm SameSize}(x,y) & ~~(3)\\
{\rm Company}(x) \rightarrow \exists z~{\rm Group}(x,z) & ~~(4) \\
{\rm Company}(x),{\rm Company}(y),{\rm Subsidiary}(x,y),{\rm Group}(y,z) \rightarrow {\rm Group}(x,z) & ~~(5)\\
0.7 :: {\rm Company}(x),{\rm Company}(y),{\rm Group}(x,z),{\rm Group}(y,z), 
{\rm Industry}(x,w),{\rm Industry}(y,w) \rightarrow \\{\rm SameSize}(x,y) & ~~(6)
\end{align*}
\label{ex:recordlinkage}
\vspace{-3mm}
\end{example}
Rules~(1) and (2) \change{increase the matching probability on the basis of common industry and comparable size, two features that we actually verified to be selective in this respect: especially in small markets, companies of the same size active in the same economic area tend to be small clusters.}
Rule~(3) defines when the same size can be assumed for $x$ and $y$ on the basis of an absolute maximum deviation $\epsilon$. Hard Rules~(4) and~(5) establish a transitive relation of the condition of ``being part of a group''; in particular, assumed that every company is within a group, the singleton one as a limit case, whenever $x$ is a subsidiary of $y$, then $x$ inherits the groups from $y$. Finally, Rule~(6) establishes a probability for two companies of being of the same size, whenever they are part of the same group and operate in the same industry, \change{based on a large body of supporting data.}

\subsection{Data Fusion}
\label{sec:data-fusion} 

Data fusion addresses the challenge of merging the facts of the same real-world entity into one single fact~\cite{BlNa08}. To achieve this goal, data fusion is concerned with solving attribute-level conflicts that can originate from disagreeing or poor quality sources and schema-level heterogeneity. 
Most of the techniques that have been proposed~\cite{YiHY08,BSDM09,DoBS15} adopt a ``truth discovery approach'' and perform metadata- and instance-based conflict resolution. 
In this section we show an example where probabilistic knowledge graphs are effectively used to model a data fusion setting where multiple and mutually dependent sources need to be harmonized. The use of PKGs generalizes early SRL approaches to data fusion, e.g., with Bayesian networks~\cite{Laur15}.

\begin{figure}
\begin{center}
\includegraphics[scale=0.4]{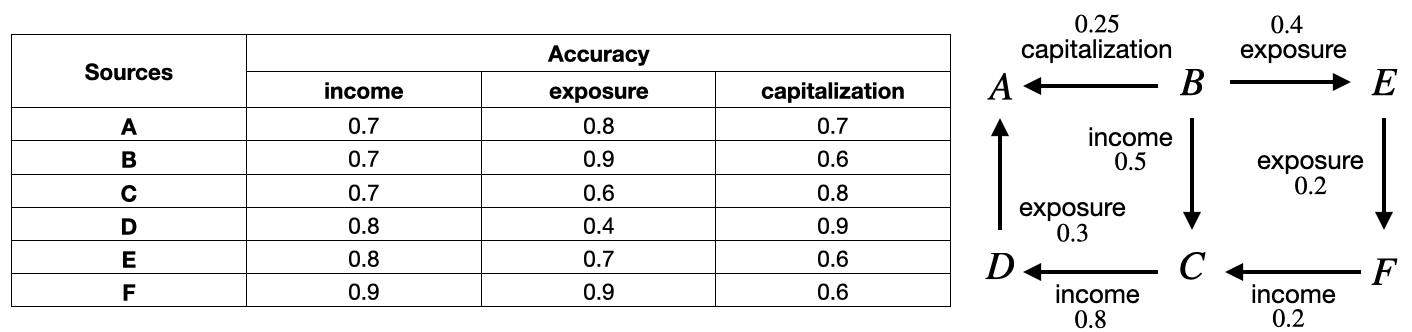} 
\caption{A data fusion use case: source accuracy by feature and influence graph.}
\end{center}
\label{fig:data_fusion_f}
\end{figure}

\smallskip\noindent
Consider the data fusion use case shown in Figure~2 about six financial data providers (A-F), providing three indicators (income, exposure and capitalization) about companies.

\begin{example}
\label{ex:data_fusion}
\textit{For each company, data providers express a value, encoded by a fact $\textsf{Vote}({\rm source},{\rm com}\-{\rm pany},\-{\rm feature},{\rm value})$, in our PKG. Providers $s$ have different levels $\alpha^s_f$ of accuracy for each feature $f$, or, in other terms, $1-\alpha^s_f$ is the error rate of $s$ for feature $f$. This value is given a priori, on the basis of the experts' knowledge and trust in the source for specific data. In our PKG, we express the accuracy of each source with a ground rule of the form $\alpha :: \textsf{Accuracy}({\rm source},{\rm feature})$. Moreover, sources are not independent, but copy from one another with a given probability. \change{This is} expressed in the dependency graph in Figure~2, where edges stand for the ``copied by'' relationship and are labelled with the feature and respective copy likelihood $\lambda$. The copy relationships are modeled with ground rules of the form: $\lambda :: \textsf{Copies}({\rm source},{\rm target},{\rm feature})$. Then, the following hard rules allow to take decisions on conflicting values.}
\change{\begin{align*}
{\rm Copies}(s,u,f) \to {\rm doesCopy}(s,f) & ~~~~(1) \\
{\rm Vote}(s,c,f,v),~\mathit{not}~{\rm doesCopy}(s,f), {\rm Accuracy}(s,f) \to {\rm Value}(c,f,v) & ~~~~(2)\\
{\rm Copies}(x,z,f), {\rm Copies}(z,y,f) \to {\rm Copies}(x,y,f) & ~~~~(3)
\end{align*}\vspace{-3mm}}
\end{example}
According to Rule~(2), a vote expressed by a source $s$ for a value $v$ of for the feature $f$ of company $c$ will turn into a fact for \textsf{Value} with a probability that is positively affected by the accuracy of $s$ on $f$ and negatively by the fact that $s$ copies such value from some other source. \change{Rule~(1) accounts for the features $f$ for which $s$ is a copier.} \change{The negation used here uses stratified semantics, as we have seen in Section~\ref{sec:kgs}.}
Rule~(3) uses recursion to model the propagation of copies in the influence graph and so affect the marginal probability of facts for \textsf{Value} accordingly.
In total, the marginal probabilities of each single \textsf{Value} represents the likelihood of choosing $v$ as a value to solve conflicts on $f$. More precisely, for each pair $\langle c,f \rangle$, the value $v$ corresponding to the fact with the maximum likelihood will be the solution to the conflict posed by the data fusion setting.

\section{Implementation and Experiments}
\label{sec:experiments}

We implemented the \svadalog framework for probabilistic reasoning on KGs as an extension of the \vadalog system. 
Coherently with the overall architecture of the systems, for the MC\-MC-chase we adopted a \emph{pipeline approach}. A set of \svadalog rules is compiled into an execution pipeline, where each pipeline node implements a rule, and nodes are connected if there is a dependency between rules.
Data flow along the pipeline, from source nodes, the EDBs, towards the target node, which corresponds to the \textsf{Ans} atom of the reasoning task. The execution pipeline undergoes an \textit{optimization phase}, where common query answering heuristics are applied.
At runtime, we adopt a \textit{pull-based approach}: the target node actively polls its predecessors in the pipeline for new facts, and so recursively the polling reaches the source nodes, feeding the pipeline with EDB facts. 
The isomorphism check applied in the warded chase steps is implemented in the form of a \textit{termination strategy}, a component that filters the fact originating from the pipeline nodes and guarantees termination of the process. 

MCMC-chase is implemented with two main techniques: a \textit{routing strategy} and an \textit{edge filter}. The routing strategy decides the number and order of activation of the pipeline nodes (lines~9 and 12 of Algorithm~\ref{alg:mcmc_chase}). The edge filters, sample the facts moving from one node to the other, choosing those to be actually propagated according to the weight of the rules they represent (and so implementing rules 10-11). A \textit{probability manager} handles the partitioning of facts into nodes of the chase network and takes decisions on \change{their} acceptability (line~13); it handles a set of supplementary buffer caches, allowing to revert rules that need to be \change{rolled back.} \change{A shared in-memory structure, the \textit{chase graph}, allows to efficiently identify applicable and undoable rules by tracing the provenance of each fact in terms of applied rule and unifier.}

\subsection{Experimental Settings}

We tested our approach using a total of 16 real-world and synthetic KGs. For the \emph{real-world} KGs, we used 4 graphs of increasing density derived from the real graph of \emph{European Financial Companies}, \change{by considering increasingly broader notions of company ownership.} The details of the topology 
are shown in Figure~\ref{tab:bdi} (upper part). For the \emph{synthetic} KGs, we built 12 \change{scale-free graphs, i.e., the degree distribution $P(k)$ of nodes with degree $k$, goes as $k^{-\gamma}$, where the parameter $\gamma$ is such that $2<\gamma<3$~\cite{BBCR03,HiBa08}.} 
Scale-free networks can be shaped via four parameters:  \(n\), the expected number of nodes; \(\alpha\), probability of adding a new node connected to an existing one; \(\beta\), and \(\gamma\), probabilities of adding an edge between two existing nodes, randomly chosen from the in-degree or out-degree distribution, respectively. We used three graph topologies, \change{denser than the real-world ones,} (namely, \change{\emph{BASE},} \emph{DENSE} and \emph{SUPER-DENSE}), in  Figure~\ref{tab:bdi} (lower part) and we varied the number of nodes from 100 to 1K.

\begin{figure}
\centering

\begin{minipage}{.5\linewidth}
\scalebox{0.8}{
\begin{tabular}{ccccc}
\multicolumn{5}{c}{\textbf{Real-world KGs}} \\
{\bf KG} & {\bf nodes} & {\bf edges} & {\bf nodes in max cc} & {\bf edges in max cc}\\
\emph{SCALE-1} & 70316 & 44716  & \(\sim\)4K & \(\sim\)5K\\
\emph{SCALE-2} & 70324 & 44721 & \(\sim\)5K & \(\sim\)6K\\
\emph{SCALE-3} & 70328 & 44722 & \(\sim\)6K & \(\sim\)8K\\
\emph{SCALE-4} & 70328 & 46408 & \(\sim\)12K & \(\sim\)13K\\
\end{tabular}}
\end{minipage}
\begin{minipage}{.3\linewidth}
\scalebox{0.8}{
\begin{tabular}{cccc}
\multicolumn{4}{c}{\textbf{Synthetic KGs}} \\
{\bf KG} & {\bf \(\alpha\)} & {\bf \(\beta\)} & {\bf \(\gamma\)}\\
\emph{\change{BASE}} & 0.71 & 0.09 & 0.2\\
\emph{DENSE} & 0.51 & 0.34 & 0.15\\
\emph{SUPER-DENSE} & 0.51 & 0.44 & 0.05\\
\multicolumn{4}{c}{\(n \in \{100, 250, 500, 1K\}\)}
\end{tabular}}
\end{minipage}

\caption{Characteristics of the real-world and synthetic KGs.}
\label{tab:bdi}
\end{figure}
 
\smallskip\noindent\textbf{The \svadalog Rules}. Our KGs are augmented with a \svadalog program 
describing the domain. A company \(x\) controls a company \(y\) if either
of the following holds: (i) \ \(x\) owns more than 50\% of the shares of \(y\); 
(ii) \(x\) controls a set of companies, which jointly, and possibly together with \(x\) direct possession, own more than
50\% of \(y\). Uncertainty can depend on three causes: relations present in the EDB not existing in reality; invalid shares in direct ownerships (greater than one or less than zero); invalid shares in indirect ownerships (greater than one).

\begin{example}
\label{ex:experiment} \textit{We model the three sources of uncertainty and assign specific weights
to rules, normalizing them between zero and one as follows.}
{
\begin{align*}
0.9 :: {\rm InputOwn}(x,y,s), 0<s<1  \rightarrow {\rm Own}(x,y,s) & ~~(1)  \\
0.1:: {\rm InputOwn}(x,y,s), (s<0\vee s>1) \rightarrow \exists z~{\rm Own}(x,y,z),{\rm Unreliable}(x,y) & ~~(2)\\
{\rm Own}(x,y,s), ~\textit{not}~{\rm Unreliable}(x,y), s > 0.5 \rightarrow {\rm Control}(x,y)& ~~(3) \\
0.5::~{\rm Own}(x,y,s), {\rm Unreliable}(x,y) \rightarrow {\rm Control}(x,y) & ~~(4)\\
{\rm Control}(x,y), {\rm Own}(y,z,s),~\textit{not}~{\rm Unreliable}(y,z), v = \textrm{sum}(s), v > 0.5 \rightarrow {\rm Control}(x,z) & ~~(5)\\
0.3::{\rm Control}(x,y), {\rm Own}(y,z,s), {\rm Unreliable}(y,z) \rightarrow {\rm Control}(x,z). & ~~(6) \\
{\rm Company}(x) \to {\rm Control}(x,x). &~~(7)
\end{align*}
}

\noindent
Rule (1) extracts valid relationships. The high weight witnesses a \change{10\%} error rate in the original data source. Rule (2) extracts invalid relationships, which, with some low probability, correspond to actually existing shares, whose amount is replaced with labelled nulls; they are marked as ``unreliable''.
Rules~(3) and~(4) handle direct control in the unreliable and reliable case, respectively. Unreliable ownerships produce control with 50\% probability, as the share amount is actually unknown.
Rule (5) extends control with a reliable ownership, generating a reliable control.
Rule (6) extends control in recursive cases with unreliable ownerships. \change{As by Rule~(7) every company controls itself, Rules~(5) and~(6) also consider direct possession.}   
In Rule~(5), the \textsf{sum} function denotes an aggregate summation operator, that accumulates the values $s$ for \textsf{Own}.

\label{ex:exp}
\end{example}
\smallskip\noindent \textbf{Settings and metrics.}  For each KG we considered the reasoning task consisting in querying the \textsf{Control} relation, i.e., enumerating all
the pairs of controller-controlled companies.

\smallskip\noindent\underline{Full grounding and exact inference}: We calculated the full grounding of the chase network with the \vadalog system and exact marginal inference by exhaustively exploring all the nodes. 

\smallskip\noindent\underline{MCMC-chase}: 
We compared the MCMC-chase settings with the exact inference on: (i) \emph{execution time}: the time (averaged over 5 executions) needed to run a predefined number of iterations 
in the MCMC-chase vs.\ the time needed for exact inference. A three-hour time-out was considered and we aborted exceeding executions; (ii) \emph{error rate}: the percentage difference (averaged over 5 executions and all the facts) between the marginal probability of the facts generated by the MCMC-chase and the marginal probability of the facts generated by exact inference;
(iii) \emph{acceptance rate}: the fraction of explored possible worlds that are accepted throughout the sampling.
For the MCMC-chase, we applied an increasing number of iterations, proportional to the number of facts in the
EDB of each KG, so that \(N_i\) denotes as many iterations as \(i\) times the number of facts. This empiric choice proved to be effective to highlight the entire error rate spectrum.

\begin{figure*}
\centering
\includegraphics[scale=0.23]{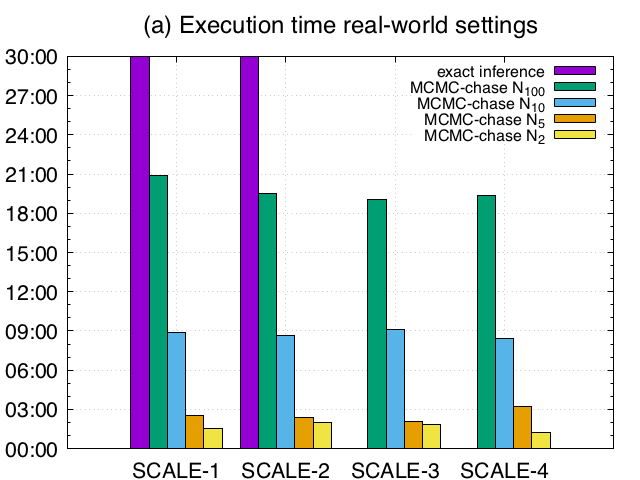}\includegraphics[scale=0.23]{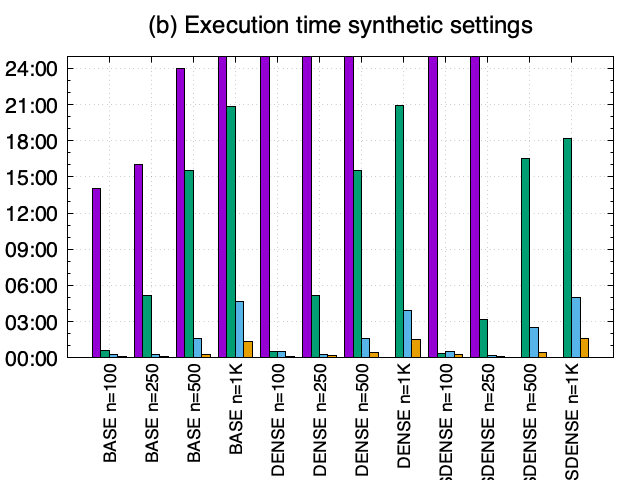}
\includegraphics[scale=0.23]{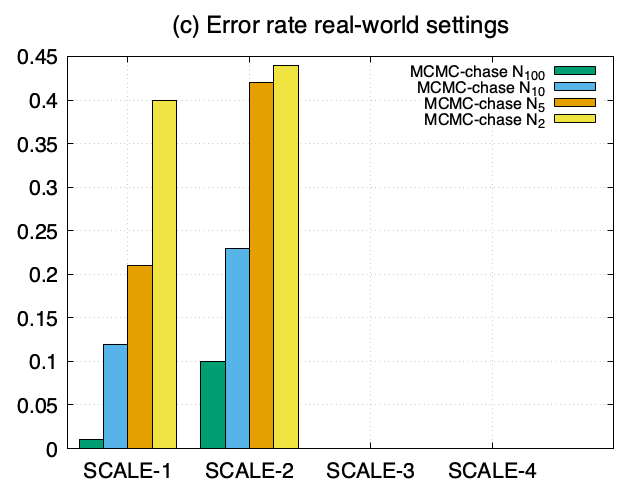}\includegraphics[scale=0.23]{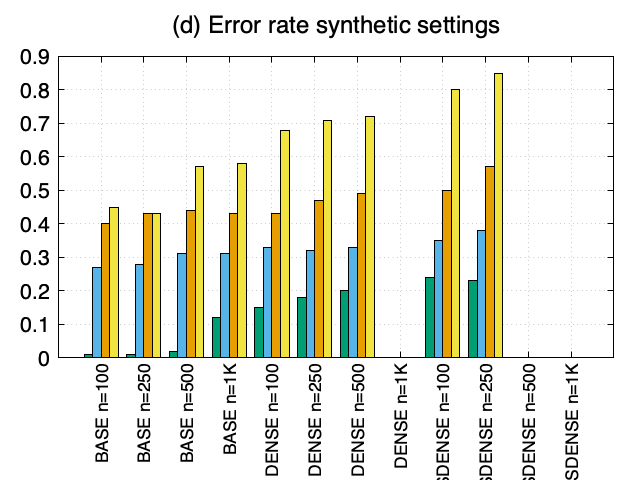}
\caption{Execution times and error rates for real-world (a-c) and synthetic (b-d) settings.}
\label{fig:experiments}
\end{figure*}

\smallskip \noindent \textbf{Results.} In real-world settings, in Figure~\ref{fig:experiments}(a-c), 
exact inference did not exceed the timeout only for \emph{SCALE-1} and \emph{SCALE-2},
which completed in $02{\rm :}09{\rm :}53$ and $02{\rm :}30{\rm :}32$, respectively.
This confirms that exact inference is not affordable in most real cases.
MCMC-chase with $N_{100}$ outperformed exact inference, completing in $\sim$ 21 minutes, with an error rate less than $2\%$. 
\change{Clearly, the MCMC approach allows to keep the elapsed time only dependent on the number of iterations.}
In fact, for $N_{100}$, the elapsed time is stable, between $18$ and $21$
minutes also for \emph{SCALE-2}, \emph{SCALE-3} and \emph{SCALE-4}. The same behaviour
can be observed for the other configurations $N_i$.
As topologies get denser, we need more iterations: \change{the complete graph is the most dense and therefore the hardest topology to sample.} This can be observed with \emph{SCALE-2},
where for $N_{100}$ error rate rises to $10\%$. Configuration $N_{10}$
takes less than $9$ minutes, with $12\%$ and \change{$23\%$} error rate, respectively for 
\emph{SCALE-1} and \emph{SCALE-2}. Observe that MCMC-chase also completes
in very short time for \emph{SCALE-3} and \emph{SCALE-4}, though
error rate cannot be calculated, because we could not calculate the exact inference baseline. 
The observed average acceptance rate is \(83.43\%\), hence fully satisfactory. 
Results in Figure~\ref{fig:experiments}(a,c) have been obtained \change{by}
fixing \(\lambda=5\) as a parameter for the jump distribution.
Interestingly, for executions
with high number of iterations, higher values for \(\lambda\)
tend to produce higher elapsed times and \change{a} smaller acceptance rate, while error rate
is stable. For example, we observed 
\(78.22\%\) acceptance rate for \(\lambda=7\) in \emph{SCALE-1} for \(N_{100}\),
with more than 30 minutes elapsed. Smaller values of \(\lambda\) reduce elapsed time and do not improve acceptance rate. 

Synthetic settings, Figure~\ref{fig:experiments}(b,d), \change{are more time-intensive than the real cases, \changeR{due to} the higher density of the graphs.} In particular, exact inference exceeded
timeout for \emph{DENSE} with $n=1K$ and for \emph{SUPER-DENSE} with $n=500$
and $n=1K$. For KGs with small $n$, independently 
of the topology, MCMC-chase is extremely performant and accurate, e.g., for \emph{SUPER-DENSE}
with $n=250$ it terminates in $\sim 3$ minutes (vs $02{\rm :}35{\rm :}20$) for $N_{100}$ with error rate
$<3\%$. Also in these settings, less dense topologies require less iterations: for example
$N_{100}$ for \emph{DENSE} with $n=250$ achieves $<2\%$ error rate, and 
\emph{\change{BASE}}even $1\%$. The acceptance rate was satisfactory (81\% on average) and we observed  similar variations as in real-world cases when adjusting $\lambda$.\footnote{\change{The artificial datasets, the execution times, and error rates are available online (\url{https://bit.ly/3IK2ooy}). The Vadalog system can be made available to research partners upon request for non-commercial use.}} 

\section{Conclusion}
\label{sec:conclusion}
\change{A probabilistic extension of Warded Datalog$^\pm$ enables effective uncertain reasoning on Knowledge Graphs, with the possibility of encoding complex domains of interest.} The probabilistic toolbox available in the literature offers insufficient support, with the impossibility to deal with existential quantification and recursion altogether. 

In this paper, we considered the reasoning desiderata for KGs and introduced the syntax and semantics of \svadalog. Within a new probabilistic reasoning framework, \svadalog allows to induce a probability distribution over the facts defined through the warded chase, a finite logical derivation procedure under which query answering is decidable and tractable. We introduced the notion of Probabilistic Knowledge Graphs, a template for chase networks, a new probabilistic graphical model where marginal inference can be performed. To cope with intractability of marginal inference, we introduced the MCMC-chase, whose core idea is performing logical and probabilistic inference at the same time, while sampling the chase space with a Monte Carlo technique. 

It is our intention to continue evolving the theoretical perspectives as well as their implementation in the system. 
\change{The probabilistic semantics we have introduced so far is coupled to the adopted chase in that multiple isomorphic copies are not considered in warded chase steps. Possible chase variants, such as other more tolerant forms of restricted and terminating chase (e.g., the Skolem chase) would potentially introduce multiple copies of isomorphic facts, arbitrarily increasing their marginal probability. In this respect, the warded semantics is the most compact and less redundant one for probabilistic reasoning and is, at the same time, sufficient for query answering, as we have seen. Along these lines, we plan to evolve the approach to adopt our isomorphism-based semantics while being chase-independent: independently of the applied chase variant, the probabilities should be computed on the isomorphism quotient set.}

\noindent
{\small
\textbf{Competing interests}. The author(s) declare none.
}

\noindent
{\small
\textbf{Acknowledgements}. The work on this paper was partially supported by the Vienna Science and Technology Fund (WWTF) grant VRG18-013.
}

\vspace{-8mm}
\bibliographystyle{acmtrans}
\bibliography{b}

\begin{thebibliography}{}

\bibitem[\protect\citeauthoryear{Agresti and Kateri}{Agresti and
  Kateri}{2011}]{AgKa11}
{\sc Agresti, A.} {\sc and} {\sc Kateri, M.} 2011.
\newblock Categorical data analysis.
\newblock In {\em International Encyclopedia of Statistical Science}. Springer,
  206--208.

\bibitem[\protect\citeauthoryear{Alberti, Bellodi, Cota, Riguzzi, and
  Zese}{Alberti et~al\mbox{.}}{2017}]{ABCR17}
{\sc Alberti, M.}, {\sc Bellodi, E.}, {\sc Cota, G.}, {\sc Riguzzi, F.}, {\sc
  and} {\sc Zese, R.} 2017.
\newblock cplint on {SWISH:} probabilistic logical inference with a web
  browser.
\newblock {\em IA\/}~{\em 11,\/}~1, 47--64.

\bibitem[\protect\citeauthoryear{Angles}{Angles}{2018}]{Angl18}
{\sc Angles, R.} 2018.
\newblock The property graph database model.
\newblock In {\em {AMW}}. Vol. 2100.

\bibitem[\protect\citeauthoryear{Bacchus}{Bacchus}{1990}]{Bacc90}
{\sc Bacchus, F.} 1990.
\newblock {\em Representing and reasoning with probabilistic knowledge - a
  logical approach to probabilities}.
\newblock {MIT} Press.

\bibitem[\protect\citeauthoryear{Beame, den Broeck, Gribkoff, and Suciu}{Beame
  et~al\mbox{.}}{2014}]{BBGS14}
{\sc Beame, P.}, {\sc den Broeck, G.~V.}, {\sc Gribkoff, E.}, {\sc and} {\sc
  Suciu, D.} 2014.
\newblock Symmetric weighted first-order model counting.
\newblock {\em CoRR\/}~{\em abs/1412.1505}.

\bibitem[\protect\citeauthoryear{Bellomarini, Fakhoury, Gottlob, and
  Sallinger}{Bellomarini et~al\mbox{.}}{2019}]{BFGS19}
{\sc Bellomarini, L.}, {\sc Fakhoury, D.}, {\sc Gottlob, G.}, {\sc and} {\sc
  Sallinger, E.} 2019.
\newblock Knowledge graphs and enterprise {AI:} the promise of an enabling
  technology.
\newblock In {\em {ICDE}}. {IEEE}, 26--37.

\bibitem[\protect\citeauthoryear{Bellomarini, Fayzrakhmanov, Gottlob,
  Kravchenko, Laurenza, Nenov, Reissfelder, Sallinger, Sherkhonov, and
  Wu}{Bellomarini et~al\mbox{.}}{2018}]{BFGK18}
{\sc Bellomarini, L.}, {\sc Fayzrakhmanov, R.~R.}, {\sc Gottlob, G.}, {\sc
  Kravchenko, A.}, {\sc Laurenza, E.}, {\sc Nenov, Y.}, {\sc Reissfelder, S.},
  {\sc Sallinger, E.}, {\sc Sherkhonov, E.}, {\sc and} {\sc Wu, L.} 2018.
\newblock Data science with {V}adalog: Bridging machine learning and reasoning.
\newblock In {\em {MEDI}}. Vol. 11163. Springer, 3--21.

\bibitem[\protect\citeauthoryear{Bellomarini, Gottlob, Pieris, and
  Sallinger}{Bellomarini et~al\mbox{.}}{2017}]{BGPS17}
{\sc Bellomarini, L.}, {\sc Gottlob, G.}, {\sc Pieris, A.}, {\sc and} {\sc
  Sallinger, E.} 2017.
\newblock Swift logic for big data and knowledge graphs.
\newblock In {\em {IJCAI}}. 2--10.

\bibitem[\protect\citeauthoryear{Bellomarini, Laurenza, Sallinger, and
  Sherkhonov}{Bellomarini et~al\mbox{.}}{2020}]{BLSS20}
{\sc Bellomarini, L.}, {\sc Laurenza, E.}, {\sc Sallinger, E.}, {\sc and} {\sc
  Sherkhonov, E.} 2020.
\newblock Reasoning under uncertainty in knowledge graphs.
\newblock In {\em RuleML+RR}. Vol. 12173. Springer, 131--139.

\bibitem[\protect\citeauthoryear{Bellomarini, Sallinger, and
  Gottlob}{Bellomarini et~al\mbox{.}}{2018}]{BeSaGo18}
{\sc Bellomarini, L.}, {\sc Sallinger, E.}, {\sc and} {\sc Gottlob, G.} 2018.
\newblock {The Vadalog System: Datalog-based Reasoning for Knowledge Graphs}.
\newblock In {\em {VLDB}}.

\bibitem[\protect\citeauthoryear{Berti{-}{\'{E}}quille, Sarma, Dong, Marian,
  and Srivastava}{Berti{-}{\'{E}}quille et~al\mbox{.}}{2009}]{BSDM09}
{\sc Berti{-}{\'{E}}quille, L.}, {\sc Sarma, A.~D.}, {\sc Dong, X.}, {\sc
  Marian, A.}, {\sc and} {\sc Srivastava, D.} 2009.
\newblock Sailing the information ocean with awareness of currents: Discovery
  and application of source dependence.
\newblock {\em CoRR\/}~{\em abs/0909.1776}.

\bibitem[\protect\citeauthoryear{Bleiholder and Naumann}{Bleiholder and
  Naumann}{2008}]{BlNa08}
{\sc Bleiholder, J.} {\sc and} {\sc Naumann, F.} 2008.
\newblock Data fusion.
\newblock {\em {ACM} Comput. Surv.\/}~{\em 41,\/}~1, 1:1--1:41.

\bibitem[\protect\citeauthoryear{Bollob\'{a}s, Borgs, Chayes, and
  Riordan}{Bollob\'{a}s et~al\mbox{.}}{2003}]{BBCR03}
{\sc Bollob\'{a}s, B.}, {\sc Borgs, C.}, {\sc Chayes, J.}, {\sc and} {\sc
  Riordan, O.} 2003.
\newblock Directed scale-free graphs.
\newblock In {\em SODA}. 132--139.

\bibitem[\protect\citeauthoryear{Borgwardt, Ceylan, and Lukasiewicz}{Borgwardt
  et~al\mbox{.}}{2017}]{BoCL17}
{\sc Borgwardt, S.}, {\sc Ceylan, {\.I}.~{\.I}.}, {\sc and} {\sc Lukasiewicz,
  T.} 2017.
\newblock Ontology-mediated queries for probabilistic databases.
\newblock In {\em {AAAI}}. {AAAI} Press, 1063--1069.

\bibitem[\protect\citeauthoryear{Borgwardt, Ceylan, and Lukasiewicz}{Borgwardt
  et~al\mbox{.}}{2018}]{BoCL18}
{\sc Borgwardt, S.}, {\sc Ceylan, {\.I}.~{\.I}.}, {\sc and} {\sc Lukasiewicz,
  T.} 2018.
\newblock Recent advances in querying probabilistic knowledge bases.
\newblock In {\em {IJCAI}}. 5420--5426.

\bibitem[\protect\citeauthoryear{Cal\`i, Gottlob, and Pieris}{Cal\`i
  et~al\mbox{.}}{2012}]{CaGP12}
{\sc Cal\`i, A.}, {\sc Gottlob, G.}, {\sc and} {\sc Pieris, A.} 2012.
\newblock Towards more expressive ontology languages: The query answering
  problem.
\newblock {\em Artif. Intell.\/}~{\em 193}, 87--128.

\bibitem[\protect\citeauthoryear{Ceri, Gottlob, Tanca, et~al\mbox{.}}{Ceri
  et~al\mbox{.}}{1989}]{ceri1989you}
{\sc Ceri, S.}, {\sc Gottlob, G.}, {\sc Tanca, L.}, {\sc et~al\mbox{.}} 1989.
\newblock What you always wanted to know about datalog (and never dared to
  ask).
\newblock {\em KDE\/}~{\em 1,\/}~1, 146--166.

\bibitem[\protect\citeauthoryear{Ceylan and Pe{\~{n}}aloza}{Ceylan and
  Pe{\~{n}}aloza}{2015}]{CePe15}
{\sc Ceylan, {\.I}.~{\.I}.} {\sc and} {\sc Pe{\~{n}}aloza, R.} 2015.
\newblock Probabilistic query answering in the bayesian description logic
  \emph{BE}l.
\newblock In {\em {SUM}}. Lecture Notes in Computer Science, vol. 9310.
  Springer, 21--35.

\bibitem[\protect\citeauthoryear{Christen}{Christen}{2012}]{Chri12}
{\sc Christen, P.} 2012.
\newblock {\em Data Matching - Concepts and Techniques for Record Linkage,
  Entity Resolution, and Duplicate Detection}.
\newblock Springer.

\bibitem[\protect\citeauthoryear{Culotta and McCallum}{Culotta and
  McCallum}{2005}]{CuMc05}
{\sc Culotta, A.} {\sc and} {\sc McCallum, A.} 2005.
\newblock Joint deduplication of multiple record types in relational data.
\newblock In {\em {CIKM}}. {ACM}, 257--258.

\bibitem[\protect\citeauthoryear{Dalvi and Suciu}{Dalvi and
  Suciu}{2007}]{DaSu07}
{\sc Dalvi, N.~N.} {\sc and} {\sc Suciu, D.} 2007.
\newblock Management of probabilistic data: foundations and challenges.
\newblock In {\em PODS}. 1--12.

\bibitem[\protect\citeauthoryear{Dalvi and Suciu}{Dalvi and
  Suciu}{2012}]{DaSu12}
{\sc Dalvi, N.~N.} {\sc and} {\sc Suciu, D.} 2012.
\newblock The dichotomy of probabilistic inference for unions of conjunctive
  queries.
\newblock {\em J. {ACM}\/}~{\em 59,\/}~6, 30:1--30:87.

\bibitem[\protect\citeauthoryear{d'Amato, Fanizzi, and Lukasiewicz}{d'Amato
  et~al\mbox{.}}{2008}]{DaFL08}
{\sc d'Amato, C.}, {\sc Fanizzi, N.}, {\sc and} {\sc Lukasiewicz, T.} 2008.
\newblock Tractable reasoning with bayesian description logics.
\newblock In {\em {SUM}}. Lecture Notes in Computer Science, vol. 5291.
  Springer, 146--159.

\bibitem[\protect\citeauthoryear{Dantsin}{Dantsin}{1991}]{Dant91}
{\sc Dantsin, E.} 1991.
\newblock Probabilistic logic programs and their semantics.
\newblock In {\em {RCLP}}. Lecture Notes in Computer Science, vol. 592.
  Springer, 152--164.

\bibitem[\protect\citeauthoryear{Dantsin, Eiter, Gottlob, and Voronkov}{Dantsin
  et~al\mbox{.}}{2001}]{DEGV01}
{\sc Dantsin, E.}, {\sc Eiter, T.}, {\sc Gottlob, G.}, {\sc and} {\sc Voronkov,
  A.} 2001.
\newblock Complexity and expressive power of logic programming.
\newblock {\em {ACM} Comput. Surv.\/}~{\em 33,\/}~3, 374--425.

\bibitem[\protect\citeauthoryear{{De Raedt} and Kimmig}{{De Raedt} and
  Kimmig}{2015}]{RaedtK15}
{\sc {De Raedt}, L.} {\sc and} {\sc Kimmig, A.} 2015.
\newblock Probabilistic (logic) programming concepts.
\newblock {\em ML\/}~{\em 100,\/}~1, 5--47.

\bibitem[\protect\citeauthoryear{den Broeck and Suciu}{den Broeck and
  Suciu}{2017}]{BrSu17}
{\sc den Broeck, G.~V.} {\sc and} {\sc Suciu, D.} 2017.
\newblock Query processing on probabilistic data: {A} survey.
\newblock {\em Found. Trends Databases\/}~{\em 7,\/}~3-4, 197--341.

\bibitem[\protect\citeauthoryear{Domingos and Lowd}{Domingos and
  Lowd}{2019}]{DoLo19}
{\sc Domingos, P.~M.} {\sc and} {\sc Lowd, D.} 2019.
\newblock Unifying logical and statistical {AI} with markov logic.
\newblock {\em {CACM}\/}~{\em 62,\/}~7, 74--83.

\bibitem[\protect\citeauthoryear{Dong, Berti{-}{\'{E}}quille, and
  Srivastava}{Dong et~al\mbox{.}}{2015}]{DoBS15}
{\sc Dong, X.~L.}, {\sc Berti{-}{\'{E}}quille, L.}, {\sc and} {\sc Srivastava,
  D.} 2015.
\newblock Data fusion: Resolving conflicts from multiple sources.
\newblock {\em CoRR\/}~{\em abs/1503.00310}.

\bibitem[\protect\citeauthoryear{Fagin, Kolaitis, Miller, and Popa}{Fagin
  et~al\mbox{.}}{2005}]{FKMP05}
{\sc Fagin, R.}, {\sc Kolaitis, P.~G.}, {\sc Miller, R.~J.}, {\sc and} {\sc
  Popa, L.} 2005.
\newblock Data exchange: semantics and query answering.
\newblock {\em Theor. Comput. Sci.\/}~{\em 336,\/}~1, 89--124.

\bibitem[\protect\citeauthoryear{Fayzrakhmanov, Sallinger, Spencer, Furche, and
  Gottlob}{Fayzrakhmanov et~al\mbox{.}}{2018}]{FSSF18}
{\sc Fayzrakhmanov, R.~R.}, {\sc Sallinger, E.}, {\sc Spencer, B.}, {\sc
  Furche, T.}, {\sc and} {\sc Gottlob, G.} 2018.
\newblock Browserless web data extraction: Challenges and opportunities.
\newblock In {\em {WWW}}. {ACM}, 1095--1104.

\bibitem[\protect\citeauthoryear{Fellegi and Sunter}{Fellegi and
  Sunter}{1969}]{FeSu69}
{\sc Fellegi, I.} {\sc and} {\sc Sunter, A.} 1969.
\newblock A theory for record linkage.
\newblock {\em Journal of American Statistical Association\/}~{\em 64},
  1183--1210.

\bibitem[\protect\citeauthoryear{Fierens, den Broeck, Renkens, Shterionov,
  Gutmann, Thon, Janssens, and Raedt}{Fierens et~al\mbox{.}}{2015}]{FiBrRe15}
{\sc Fierens, D.}, {\sc den Broeck, G.~V.}, {\sc Renkens, J.}, {\sc Shterionov,
  D.~S.}, {\sc Gutmann, B.}, {\sc Thon, I.}, {\sc Janssens, G.}, {\sc and} {\sc
  Raedt, L.~D.} 2015.
\newblock Inference and learning in probabilistic logic programs using weighted
  boolean formulas.
\newblock {\em {TPLP}\/}.

\bibitem[\protect\citeauthoryear{Gilks, Richardson, and Spiegelhalter}{Gilks
  et~al\mbox{.}}{1995}]{gilk95}
{\sc Gilks, W.}, {\sc Richardson, S.}, {\sc and} {\sc Spiegelhalter, D.} 1995.
\newblock {\em Markov Chain Monte Carlo in Practice}.
\newblock Chapman \& Hall/CRC Interdisciplinary Statistics. Taylor \& Francis.

\bibitem[\protect\citeauthoryear{Goodman, Mansinghka, Roy, Bonawitz, and
  Tenenbaum}{Goodman et~al\mbox{.}}{2008}]{GMRB08}
{\sc Goodman, N.~D.}, {\sc Mansinghka, V.~K.}, {\sc Roy, D.~M.}, {\sc Bonawitz,
  K.}, {\sc and} {\sc Tenenbaum, J.~B.} 2008.
\newblock Church: a language for generative models.
\newblock In {\em {UAI}}.

\bibitem[\protect\citeauthoryear{Gottlob, Lukasiewicz, Martinez, and
  Simari}{Gottlob et~al\mbox{.}}{2013}]{GLVS13}
{\sc Gottlob, G.}, {\sc Lukasiewicz, T.}, {\sc Martinez, M.~V.}, {\sc and} {\sc
  Simari, G.~I.} 2013.
\newblock Query answering under probabilistic uncertainty in datalog+ / -
  ontologies.
\newblock {\em Ann. Math. Artif. Intell.\/}~{\em 69,\/}~1, 37--72.

\bibitem[\protect\citeauthoryear{Gottlob and Pieris}{Gottlob and
  Pieris}{2015}]{GoPi15}
{\sc Gottlob, G.} {\sc and} {\sc Pieris, A.} 2015.
\newblock Beyond {SPARQL} under {OWL} 2 {QL} entailment regime: Rules to the
  rescue.
\newblock In {\em IJCAI}. 2999--3007.

\bibitem[\protect\citeauthoryear{Green and Tannen}{Green and
  Tannen}{2006}]{GeTa06}
{\sc Green, T.~J.} {\sc and} {\sc Tannen, V.} 2006.
\newblock Models for incomplete and probabilistic information.
\newblock {\em {IEEE} Data Eng. Bull.\/}~{\em 29,\/}~1, 17--24.

\bibitem[\protect\citeauthoryear{Gribkoff and Suciu}{Gribkoff and
  Suciu}{2016}]{GrSu16}
{\sc Gribkoff, E.} {\sc and} {\sc Suciu, D.} 2016.
\newblock Slimshot: In-database probabilistic inference for knowledge bases.
\newblock {\em {PVLDB}\/}~{\em 9,\/}~7, 552--563.

\bibitem[\protect\citeauthoryear{Halpern}{Halpern}{1989}]{Halp89}
{\sc Halpern, J.~Y.} 1989.
\newblock An analysis of first-order logics of probability.
\newblock In {\em {IJCAI}}. 1375--1381.

\bibitem[\protect\citeauthoryear{Hastings}{Hastings}{1970}]{hast70}
{\sc Hastings, W.~K.} 1970.
\newblock Monte carlo sampling methods using markov chains and their
  applications.
\newblock {\em Biometrika\/}~{\em 57,\/}~1, 97--109.

\bibitem[\protect\citeauthoryear{Hidalgo and Barab{\'{a}}si}{Hidalgo and
  Barab{\'{a}}si}{2008}]{HiBa08}
{\sc Hidalgo, C.~A.} {\sc and} {\sc Barab{\'{a}}si, A.} 2008.
\newblock Scale-free networks.
\newblock {\em Scholarpedia\/}~{\em 3,\/}~1, 1716.

\bibitem[\protect\citeauthoryear{Huang, Antova, Koch, and Olteanu}{Huang
  et~al\mbox{.}}{2009}]{HAKO09}
{\sc Huang, J.}, {\sc Antova, L.}, {\sc Koch, C.}, {\sc and} {\sc Olteanu, D.}
  2009.
\newblock Maybms: a probabilistic database management system.
\newblock In {\em SIGMOD Conference}. 1071--1074.

\bibitem[\protect\citeauthoryear{Jaeger}{Jaeger}{2018}]{Jaeg18}
{\sc Jaeger, M.} 2018.
\newblock Probabilistic logic and relational models.
\newblock In {\em Encyclopedia of Social Network Analysis and Mining. 2nd Ed.}
  Springer.

\bibitem[\protect\citeauthoryear{Jung and Lutz}{Jung and Lutz}{2012}]{JuLu12}
{\sc Jung, J.~C.} {\sc and} {\sc Lutz, C.} 2012.
\newblock Ontology-based access to probabilistic data with {OWL} {QL}.
\newblock In {\em {ISWC} {(1)}}. Lecture Notes in Computer Science, vol. 7649.
  Springer, 182--197.

\bibitem[\protect\citeauthoryear{Kersting and Raedt}{Kersting and
  Raedt}{2008}]{KeRa08}
{\sc Kersting, K.} {\sc and} {\sc Raedt, L.~D.} 2008.
\newblock Basic principles of learning bayesian logic programs.
\newblock In {\em Probabilistic Inductive Logic Programming}.

\bibitem[\protect\citeauthoryear{Koller and Friedman}{Koller and
  Friedman}{2009}]{KoFr09}
{\sc Koller, D.} {\sc and} {\sc Friedman, N.} 2009.
\newblock {\em Probabilistic Graphical Models: Principles and Techniques}.
\newblock MIT.

\bibitem[\protect\citeauthoryear{Krompa{\ss}, Nickel, and Tresp}{Krompa{\ss}
  et~al\mbox{.}}{2014}]{KrNi14}
{\sc Krompa{\ss}, D.}, {\sc Nickel, M.}, {\sc and} {\sc Tresp, V.} 2014.
\newblock Querying factorized probabilistic triple databases.
\newblock In {\em {ISWC} {(2)}}. Lecture Notes in Computer Science, vol. 8797.
  Springer, 114--129.

\bibitem[\protect\citeauthoryear{Latour, Babaki, Dries, Kimmig, den Broeck, and
  Nijssen}{Latour et~al\mbox{.}}{2017}]{LBDK17}
{\sc Latour, A. L.~D.}, {\sc Babaki, B.}, {\sc Dries, A.}, {\sc Kimmig, A.},
  {\sc den Broeck, G.~V.}, {\sc and} {\sc Nijssen, S.} 2017.
\newblock Combining stochastic constraint optimization and probabilistic
  programming - from knowledge compilation to constraint solving.
\newblock In {\em {CP}}. LNCS, vol. 10416. Springer, 495--511.

\bibitem[\protect\citeauthoryear{Laurenza}{Laurenza}{2015}]{Laur15}
{\sc Laurenza, E.} 2015.
\newblock Solving conflicts in database fusion with bayesian networks.
\newblock In {\em {FUSION}}. 399--406.

\bibitem[\protect\citeauthoryear{Lee and Wang}{Lee and Wang}{2016}]{LeWa16}
{\sc Lee, J.} {\sc and} {\sc Wang, Y.} 2016.
\newblock Weighted rules under the stable model semantics.
\newblock In {\em {KR}}. 145--154.

\bibitem[\protect\citeauthoryear{Marx, Kr{\"{o}}tzsch, and Thost}{Marx
  et~al\mbox{.}}{2017}]{MaKr17}
{\sc Marx, M.}, {\sc Kr{\"{o}}tzsch, M.}, {\sc and} {\sc Thost, V.} 2017.
\newblock Logic on {MARS:} ontologies for generalised property graphs.
\newblock In {\em {IJCAI}}. 1188--1194.

\bibitem[\protect\citeauthoryear{McCallum, Tejada, and Quass}{McCallum
  et~al\mbox{.}}{2003}]{KDD03}
{\sc McCallum, A.}, {\sc Tejada, S.}, {\sc and} {\sc Quass, D.}, Eds. 2003.
\newblock {\em Proceedings of the KDD-2003 Workshop on Data Cleaning, Record
  Linkage, and Object Consolidation}. ACM Press.

\bibitem[\protect\citeauthoryear{McCallum and Wellner}{McCallum and
  Wellner}{2004}]{McWe04}
{\sc McCallum, A.} {\sc and} {\sc Wellner, B.} 2004.
\newblock Conditional models of identity uncertainty with application to noun
  coreference.
\newblock In {\em {NIPS}}. 905--912.

\bibitem[\protect\citeauthoryear{Michels, Fayzrakhmanov, Ley, Sallinger, and
  Schenkel}{Michels et~al\mbox{.}}{2017}]{MFLS17}
{\sc Michels, C.}, {\sc Fayzrakhmanov, R.~R.}, {\sc Ley, M.}, {\sc Sallinger,
  E.}, {\sc and} {\sc Schenkel, R.} 2017.
\newblock Oxpath-based data acquisition for dblp.
\newblock In {\em {JCDL}}. {IEEE} Computer Society, 319--320.

\bibitem[\protect\citeauthoryear{Milch, Marthi, Russell, Sontag, Ong, and
  Kolobov}{Milch et~al\mbox{.}}{2005}]{MilchMRSOK05}
{\sc Milch, B.}, {\sc Marthi, B.}, {\sc Russell, S.~J.}, {\sc Sontag, D.}, {\sc
  Ong, D.~L.}, {\sc and} {\sc Kolobov, A.} 2005.
\newblock {BLOG:} probabilistic models with unknown objects.
\newblock In {\em {IJCAI}}.

\bibitem[\protect\citeauthoryear{Mumick, Pirahesh, and Ramakrishnan}{Mumick
  et~al\mbox{.}}{1990}]{MuPR90}
{\sc Mumick, I.~S.}, {\sc Pirahesh, H.}, {\sc and} {\sc Ramakrishnan, R.} 1990.
\newblock The magic of duplicates and aggregates.
\newblock In {\em VLDB} (2002-01-03), {D.~McLeod}, {R.~Sacks-Davis}, {and}
  {H.-J. Schek}, Eds. Morgan Kaufmann, 264--277.

\bibitem[\protect\citeauthoryear{Niu, R{\'{e}}, Doan, and Shavlik}{Niu
  et~al\mbox{.}}{2011}]{NRDS11}
{\sc Niu, F.}, {\sc R{\'{e}}, C.}, {\sc Doan, A.}, {\sc and} {\sc Shavlik,
  J.~W.} 2011.
\newblock Tuffy: Scaling up statistical inference in markov logic networks
  using an {RDBMS}.
\newblock {\em {PVLDB}\/}~{\em 4,\/}~6, 373--384.

\bibitem[\protect\citeauthoryear{Nocedal and Wright}{Nocedal and
  Wright}{1999}]{NoWr99}
{\sc Nocedal, J.} {\sc and} {\sc Wright, S.~J.} 1999.
\newblock {\em Numerical Optimization}.
\newblock Springer.

\bibitem[\protect\citeauthoryear{Olteanu}{Olteanu}{2016}]{Olte16}
{\sc Olteanu, D.} 2016.
\newblock Factorized databases: {A} knowledge compilation perspective.
\newblock In {\em {AAAI} Workshop: Beyond {NP}}. {AAAI} Workshops, vol.
  {WS-16-05}. {AAAI} Press.

\bibitem[\protect\citeauthoryear{Olteanu and Schleich}{Olteanu and
  Schleich}{2016}]{OlSc16}
{\sc Olteanu, D.} {\sc and} {\sc Schleich, M.} 2016.
\newblock Factorized databases.
\newblock {\em {SIGMOD} Rec.\/}~{\em 45,\/}~2, 5--16.

\bibitem[\protect\citeauthoryear{Pfeffer and River~Analytics}{Pfeffer and
  River~Analytics}{2009}]{Pfef09}
{\sc Pfeffer, A.} {\sc and} {\sc River~Analytics, C.} 2009.
\newblock Figaro: An object-oriented probabilistic programming language.

\bibitem[\protect\citeauthoryear{Poggi, Lembo, Calvanese, Giacomo, Lenzerini,
  and Rosati}{Poggi et~al\mbox{.}}{2008}]{PLCD08}
{\sc Poggi, A.}, {\sc Lembo, D.}, {\sc Calvanese, D.}, {\sc Giacomo, G.~D.},
  {\sc Lenzerini, M.}, {\sc and} {\sc Rosati, R.} 2008.
\newblock Linking data to ontologies.
\newblock {\em J. Data Semant.\/}~{\em 10}, 133--173.

\bibitem[\protect\citeauthoryear{Poole}{Poole}{1993}]{Pool93}
{\sc Poole, D.} 1993.
\newblock Logic programming, abduction and probability - {A} top-down anytime
  algorithm for estimating prior and posterior probabilities.
\newblock {\em New Gener. Comput.\/}~{\em 11,\/}~3, 377--400.

\bibitem[\protect\citeauthoryear{Poole}{Poole}{2008}]{Pool08}
{\sc Poole, D.} 2008.
\newblock The independent choice logic and beyond.
\newblock In {\em Probabilistic Inductive Logic Progr.} LNCS, vol. 4911.
  Springer, 222--243.

\bibitem[\protect\citeauthoryear{Provan and Ball}{Provan and
  Ball}{1983}]{PrBa83}
{\sc Provan, J.~S.} {\sc and} {\sc Ball, M.~O.} 1983.
\newblock The complexity of counting cuts and of computing the probability that
  a graph is connected.
\newblock {\em {SIAM} J. Comput.\/}~{\em 12,\/}~4, 777--788.

\bibitem[\protect\citeauthoryear{Richardson and Domingos}{Richardson and
  Domingos}{2006}]{RiDo06}
{\sc Richardson, M.} {\sc and} {\sc Domingos, P.~M.} 2006.
\newblock Markov logic networks.
\newblock {\em Machine Learning\/}~{\em 62,\/}~1-2, 107--136.

\bibitem[\protect\citeauthoryear{Riguzzi}{Riguzzi}{2007}]{Rigu08}
{\sc Riguzzi, F.} 2007.
\newblock A top down interpreter for {LPAD} and cp-logic.
\newblock In {\em AI*IA}. Vol. 4733. Springer, 109--120.

\bibitem[\protect\citeauthoryear{Sato}{Sato}{1995}]{Sato95}
{\sc Sato, T.} 1995.
\newblock A statistical learning method for logic programs with distribution
  semantics.
\newblock In {\em {ICLP}}. 715--729.

\bibitem[\protect\citeauthoryear{Sato and Kameya}{Sato and
  Kameya}{1997}]{SatoK97}
{\sc Sato, T.} {\sc and} {\sc Kameya, Y.} 1997.
\newblock {PRISM:} {A} language for symbolic-statistical modeling.
\newblock In {\em {IJCAI}}. 1330--1339.

\bibitem[\protect\citeauthoryear{Singla and Domingos}{Singla and
  Domingos}{2005}]{SiDo05}
{\sc Singla, P.} {\sc and} {\sc Domingos, P.~M.} 2005.
\newblock Object identification with attribute-mediated dependences.
\newblock In {\em {PKDD}}. Lecture Notes in Computer Science, vol. 3721.
  Springer, 297--308.

\bibitem[\protect\citeauthoryear{Singla and Domingos}{Singla and
  Domingos}{2006}]{SiDo06}
{\sc Singla, P.} {\sc and} {\sc Domingos, P.~M.} 2006.
\newblock Entity resolution with markov logic.
\newblock In {\em {ICDM}}. {IEEE} Computer Society, 572--582.

\bibitem[\protect\citeauthoryear{Stuart and Ord}{Stuart and
  Ord}{1991}]{Stuart1991}
{\sc Stuart, A.} {\sc and} {\sc Ord, K.} 1991.
\newblock {\em {Kendall's advanced theory of statistics}\/}, Fifth ed. Vol. 2,
  Classical Inference and Relationship.

\bibitem[\protect\citeauthoryear{Suciu, Olteanu, R{\'e}, and Koch}{Suciu
  et~al\mbox{.}}{2011}]{SORK11}
{\sc Suciu, D.}, {\sc Olteanu, D.}, {\sc R{\'e}, C.}, {\sc and} {\sc Koch, C.}
  2011.
\newblock {\em Probabilistic Databases}.
\newblock Synthesis Lectures on Data Management. Morgan {\&} Claypool
  Publishers.

\bibitem[\protect\citeauthoryear{Tierney}{Tierney}{1994}]{Tie94}
{\sc Tierney, L.} 1994.
\newblock Markov chains for exploring posterior distributions.
\newblock {\em Annals of Statistics\/}~{\em 22}, 1701--1728.

\bibitem[\protect\citeauthoryear{Ullman}{Ullman}{1997}]{Ullm97}
{\sc Ullman, J.~D.} 1997.
\newblock Information integration using logical views.
\newblock In {\em ICDT}. 19--40.

\bibitem[\protect\citeauthoryear{Vennekens, Denecker, and Bruynooghe}{Vennekens
  et~al\mbox{.}}{2009}]{Venn09}
{\sc Vennekens, J.}, {\sc Denecker, M.}, {\sc and} {\sc Bruynooghe, M.} 2009.
\newblock Cp-logic: {A} language of causal probabilistic events and its
  relation to logic programming.
\newblock {\em Theory Pract. Log. Program.\/}~{\em 9,\/}~3, 245--308.

\bibitem[\protect\citeauthoryear{Vennekens, Verbaeten, and
  Bruynooghe}{Vennekens et~al\mbox{.}}{2004}]{VeVB04}
{\sc Vennekens, J.}, {\sc Verbaeten, S.}, {\sc and} {\sc Bruynooghe, M.} 2004.
\newblock Logic programs with annotated disjunctions.
\newblock In {\em {ICLP}}.

\bibitem[\protect\citeauthoryear{Yin, Han, and Yu}{Yin
  et~al\mbox{.}}{2008}]{YiHY08}
{\sc Yin, X.}, {\sc Han, J.}, {\sc and} {\sc Yu, P.~S.} 2008.
\newblock Truth discovery with multiple conflicting information providers on
  the web.
\newblock {\em {IEEE} Trans. Knowl. Data Eng.\/}~{\em 20,\/}~6, 796--808.

\end{thebibliography}

\label{lastpage}
\end{document}